\documentclass[preprint,12pt]{elsarticle}
\PassOptionsToPackage{square,numbers,sort&compress}{natbib}

\usepackage[square,numbers,sort&compress]{natbib}   % only once

\usepackage{hyperref}  % or \usepackage{url}

\usepackage{amssymb}
\usepackage{amsmath}
\usepackage{lineno}
\usepackage{subcaption}
\usepackage{pifont}
\usepackage{makecell} % add this in your preamble
\journal{Computerized Medical Imaging and Graphics}

\begin{document}

\begin{frontmatter}

\title{\textbf{PULSE: A Unified Multi-Task Architecture for Cardiac Segmentation, Diagnosis, and Few-Shot Cross-Modality Clinical Adaptation}} %% Article title

% \author[]{Anonymous}

\author[1]{Hania Ghouse\fnref{fn1}}
\author[2]{Maryam Alsharqi\fnref{fn2}}
\author[2,3]{Farhad R. Nezami\fnref{fn3}}
\author[1,4]{Muzammil Behzad\corref{cor1}}

% ------------------ CORRESPONDING AUTHOR ------------------
\cortext[cor1]{Corresponding author: Muzammil Behzad (email: muzammil.behzad@kfupm.edu.sa)}

% ------------------------- AFFILIATIONS -------------------
\address[1]{King Fahd University of Petroleum and Minerals, Saudi Arabia}
\address[2]{Institute for Medical Engineering \& Science, Massachusetts Institute of Technology, US}
\address[3]{Harvard Medical School, Harvard University, US}
\address[4]{KFUPM–SDAIA Joint Research Centre for Artificial Intelligence, Saudi Arabia}

% ---------------------- FOOTNOTE EMAIL --------------------
\fntext[fn1]{Hania Ghouse: g202518690@kfupm.edu.sa}
\fntext[fn2]{Maryam Alsharqi: maryam7@mit.edu}
\fntext[fn3]{Farhad Nezami: frikhtegarnezami@bwh.harvard.edu}

\begin{abstract}
Cardiac image analysis remains fragmented across tasks: anatomical segmentation, disease classification, and grounded clinical report generation are typically handled by separate networks trained under different data regimes. No existing framework unifies these objectives within a single architecture while retaining generalization across imaging modalities and datasets. We introduce PULSE, a multi-task vision–language framework built on self-supervised representations and optimized through a composite supervision strategy that balances region overlap learning, pixel wise classification fidelity, and boundary aware IoU refinement. A multi-scale token reconstruction decoder enables anatomical segmentation, while shared global representations support disease classification and clinically grounded text output allowing the model to transition from pixels to structures and finally clinical reasoning within one architecture. Unlike prior task-specific pipelines, PULSE learns task-invariant cardiac priors, generalizes robustly across datasets, and can be adapted to new imaging modalities with minimal supervision. This moves the field closer to a scalable, foundation style cardiac analysis framework. 
% The project repo and codes are available at: 
% \href{https://github.com/anonymous}{GitHub Repository Link}

\end{abstract}


\end{frontmatter}

\section{Introduction}

Cardiovascular disease remains the leading cause of mortality worldwide \cite{DiCesare2024}. A core part of its clinical evaluation rests on accurate assessment of the left ventricle (LV), the heart’s main pumping chamber, because LV volumes, myocardial mass, and ejection fraction (EF) are fundamental indicators of pump function, remodeling, and disease severity \cite{12, SanRomn2009}. Cardiac magnetic resonance (CMR) imaging is the clinical standard for quantifying ventricular volumes and myocardial thickness, thanks to its superior soft-tissue contrast and reproducibility \cite{Armstrong2012}; yet segmentation still relies heavily on manual contouring, which is time-consuming, operator-dependent, and hard to scale in high-volume settings \cite{Schilling2024}.  Deep learning (DL) has emerged as the dominant approach for automating cardiac image segmentation, outperforming traditional handcrafted or atlas-based techniques across MRI, CT, and ultrasound modalities \cite{Chen2020}. Both Convolutional Neural Networks(CNNs) and newer transformer based architectures now achieve high Dice on benchmark datasets such as the ACDC challenge, reliably delineating ventricular structures even in pathological cases \cite{Bernard2018, Abdeltawab2020}. Many pipelines then compute volumetric and functional indices from these masks.   However, segmentation alone is insufficient for clinical decision support. Most DL models are developed for a single dataset, typically short-axis MRI cine stacks, under full supervision and do not generalize well across centers, scanner types, vendors, ormodalities, limiting clinical scalability \cite{Chai2025}. 

Independent research efforts performing cardiac disease classification demonstrate that while diagnostic accuracy can be high, they often rely on reduced representations (e.g., global latent vectors, volumes) and discard pixel-wise anatomy provided by segmentation masks \cite{Srikijkasemwat2025}. As a result, anatomical segmentation and disease reasoning remain decoupled: one model “sees” structure, another “sees” disease, and the connection between them is mediated only by ad-hoc feature engineering.   Real-world clinical workflows demand more: clinicians want a coherent chain from anatomy to quantitative indices to diagnosis and finally to structured report. Yet to our best knowledge, no widely adopted DL framework simultaneously delivers (a) high-fidelity segmentation, (b) disease or functional classification, and (c) clinically relevant outputs (volumetric indices or structured narrative) in a unified pipeline. Even recent cardiac segmentation studies typically conclude at mask generation \cite{Abdeltawab2020}. Although some multi-task networks attempt to combine segmentation with classification or motion estimation \cite{Peng2021}, they remain limited: most implement only two tasks (e.g. segmentation + classification), rely on CNN based backbones with limited long range context, and do not produce quantitative indices or structured clinical output . From an architectural perspective, the segmentation only paradigm also overlooks important trade-offs. 

Pure 2D slice-by-slice networks are computationally efficient and data-light, but neglect through plane context risking inconsistent volumetric indices when slice alignment, thickness, or appearance vary between acquisitions \cite{cardiacsurvey}. Fully 3D convolutional models address spatial consistency but are memory and computationally heavy, and struggle when data are anisotropic or scarce \cite{Ilesanmi2024}. Some studies therefore adopt a 2.5D compromise, processing stacks of adjacent slices together to preserve anatomical continuity while maintaining tractable computation \cite{Zhang2022}. However, such architectures are rarely extended to multi-task reasoning, diagnosis, or report generation.   These observations expose key gaps in current practice. First, task design remains fragmented: segmentation, classification, and clinical reporting are handled by separate modules or models. Second, there is little systematic study of how design choices, such as normalization strategy, loss weighting, or architecture depth, influence clinically meaningful outputs like ejection fraction error, misclassification rate, or report fidelity. Third, there is a lack of reproducible end-to-end pipelines: few works trace how improvements in segmentation propagate to diagnosis and clinical indices under varying imaging conditions.  In this study, we propose PULSE: a unified transformer-based framework that performs ventricular segmentation, cardiomyopathy classification, and structured clinical report generation directly from short-axis CMR slices. This approach treats cardiac image interpretation as a cohesive, multi-task reasoning problem, from pixels to quantitative indices to narrative summaries.

\section{Literature Review}

Deep learning (DL) has emerged as the standard for cardiac image segmentation, with fully convolutional and U-Net derived architectures remaining dominant, and newer attention- or hybrid-based models increasingly explored for richer anatomical context \cite{ElTaraboulsi2023, Anusha_Prasad_2024}. Early U-Net style methods achieved accurate delineation of LV, RV, and myocardium across modalities, as summarized in major reviews \cite{Chen2020}, and benchmark datasets such as ACDC demonstrated near-expert segmentation with clinically useful extraction of volumes and ejection fraction \cite{Bernard2018}. Despite high performance under controlled conditions, many pipelines derive functional indices via downstream post-processing rather than optimizing segmentation and metrics jointly \cite{Galati2022, Abdeltawab2020}. Some works compute volumes or EF directly from predicted masks, which reduces clinician burden but still treats segmentation and downstream quantification as separate stages \cite{Guo2022}. To address this, a few studies have adopted multi-task learning (MTL), coupling segmentation with auxiliary tasks such as cardiac phase detection, wall thickness estimation, or disease classification \cite{Vesal2021, Peng2021}. However, MTL remains rare, in part due to sensitivity to loss weighting and lack of evaluation on downstream clinical outputs. 

Another major limitation is domain shift: models trained on data from a single center or scanner often generalize poorly to unseen vendors, protocols, or pathology distributions \cite{Ugurlu2022, Patil2023}. The multi-centre, multi-vendor, multi-disease M\&Ms Challenge highlighted this problem, showing substantial performance drops when methods are evaluated on heterogeneous CMR datasets \cite{Campello2021}. Beyond segmentation, deep models must also handle variable acquisition settings (slice thickness, spatial resolution) and limited annotated data issues that hamper robust functional and volumetric quantification across cohorts \cite{Alnasser2024}.Multi-task networks continue to evolve: a 2022 MTL-UNet augmented segmentation with an edge-detection branch to improve anatomical boundaries \cite{Ren2022}. Others combine segmentation of myocardium with scar detection in infarcted tissue, showing that segmentation based features can directly aid pathology classification \cite{Xing2023}. Large-scale, multicenter pipelines have demonstrated that DL segmentation + volumetric quantification can work robustly even in heterogeneous patient populations with variable anatomy (e.g. a registry of single ventricle hearts) \cite{Yao2024}.  

Beyond cine MRI, DL segmentation is applied to 4D flow MRI for automated LV segmentation and hemodynamic quantification, avoiding need for separate cine acquisitions \cite{Sun2024}. New multiscale architectures such as DeSPPNet improve delineation of cardiac chambers under challenging image conditions \cite{Elizar2024}. Robust recent models, such as ResST-SEUNet++, have demonstrated high IoU (93.7\%) and strong generalization across multiple datasets \cite{BaMahel2025}. Although there is growing interest in self-supervised or foundation model approaches using large unlabeled datasets, to our knowledge no existing work combines such backbones with a unified multi task pipeline for segmentation, disease classification, and clinical-metric regression under cross-dataset evaluation \cite{dinov2}.  

\section{Methodology}

\subsection{Data Preprocessing and Augmentation}

A robust preprocessing pipeline is essential for ensuring stability under heterogeneous acquisition conditions. 
All datasets are converted into normalized short-axis cine volumes, followed by a structured multi-stage preparation 
protocol consisting of volume normalization, 2.5D contextual slice construction, domain-informed augmentation, 
and test-time inference correction. The objective is to preserve high-value anatomical signal while improving 
resilience to scanner variability, image contrast differences, speckle noise, and apical-basal visibility gradients.

% ------------------------ 3.1.1 ------------------------
\subsubsection{Volume-wise Normalization}

Unlike slice-wise normalization, which may amplify noise in apical or low-signal slices, we employ global 
volume-wise Z-score normalization. For a 3D cine MRI volume 
$V \in \mathbb{R}^{H \times W \times D}$ with $N = HWD$ voxels, the global mean $\mu_V$ and standard deviation 
$\sigma_V$ are computed as:

\[
\mu_V = \frac{1}{N} \sum_{i=1}^{N} V_i, \qquad
\sigma_V = \sqrt{\frac{1}{N} \sum_{i=1}^{N} (V_i - \mu_V)^2 }
\]Each voxel is standardized using:

\[
\hat{v}_{x,y,z} = \frac{v_{x,y,z} - \mu_V}{\sigma_V + \varepsilon}, \qquad \varepsilon = 10^{-6}
\]This preserves global contrast differences across basal, mid, and apical slices, strengthening myocardium-to-cavity 
intensity gradients and yielding more stable downstream feature embeddings.

% ------------------------ 3.1.2 ------------------------
\subsubsection{2.5D Context Stack Generation}

To balance volumetric anatomical context with the efficiency of 2D convolution, we adopt a 2.5D contextual input 
formulation. For slice index $z$, the network input is constructed as:

\[
X_{z} = I_{z-1} \,\oplus\, I_{z} \,\oplus\, I_{z+1} \in \mathbb{R}^{3 \times 224 \times 224}
\]
Boundary slices are padded using clamped replication:

\[
I_{z'} = I_{\max(0,\min(D-1,z'))}
\] Image slices are resized to $224{\times}224$ using bilinear interpolation, while ground-truth masks are upsampled 
with nearest-neighbor interpolation to avoid label softening. This approach retains cross-slice ventricular continuity 
without the computational overhead of full 3D networks.

% ------------------------ 3.1.3 ------------------------
\subsubsection{Domain-Robust Augmentation Strategy}

We implement a domain generalizing augmentation pipeline using Albumentations. The augmentation space models are 
clinically realistic variation, including breath-hold shift, slice-plane rotation, vendor-dependent contrast, 
and speckle-rich signal degradation. During training, each slice may undergo random spatial perturbations including in-plane rotation 
($\theta\sim U[-30^\circ, 30^\circ], p=0.5$) and shift–scale–rotate transformations with up to 10\% translation 
and 20\% isotropic scaling ($p=0.5$). Non-rigid motion variations are simulated using elastic deformation 
($\alpha=1,\sigma=50,\alpha_{\text{aff}}=50$, $p=0.3$), while grid distortion is introduced with probability $0.3$ 
to mimic slice warping and breath-hold irregularity. Horizontal and vertical flips are applied independently 
($p=0.5$ each), and intensity-space variability is modeled using additive Gaussian noise 
$\mathcal{N}(0,\sigma^{2})$ with probability $0.2$. Collectively, these operations reflect the distribution of 
real world cardiac MRI, improving model robustness against anatomical variation, scanner domain shift, and noise-induced degradation. This considerably improves the network's tolerance to intensity drift, through plane misalignment, and domain shift.

% ------------------------ 3.1.4 ------------------------
\subsubsection{Test Time Augmentation (TTA)}

Inference is stabilized using three-view test-time augmentation. For input $X$, the final prediction is:\[
P_{\text{final}}(X) = \frac{1}{3} \Big(
P(X) 
+ \text{Flip}_H^{-1}(P(\text{Flip}_H(X))) 
+ \text{Flip}_V^{-1}(P(\text{Flip}_V(X)))
\Big)
\]where inverse flips restore predictions to their native spatial frame. This reduces structural fragmentation and 
mitigates boundary uncertainty in apical and basal slices.

\subsection{ Model Framework}

The proposed framework, PULSE, is an end-to-end hybrid architecture that unifies 
pixel level anatomical segmentation, disease level classification, and transformer based 
representation learning. The core design philosophy is to capitalize on the semantic strength 
of large-scale self-supervised Vision Transformers while restoring spatial detail through a 
multiscale pyramid decoder optimized for cardiac morphology. A high-level overview of the system 
is shown in Figure~\ref{fig:architecture}, and the following subsections describe each 
module in detail.

% ---------------------- 3.2.1 ----------------------
\subsubsection{Self-Supervised Backbone (DINOv2)}

The encoder of PULSE is initialized using DINOv2 ViT-B/14, a self-supervised pretrained transformer, which yields strong feature invariance and cross-domain transferability. 
Given a normalized 2.5D slice input $X \in \mathbb{R}^{H \times W \times C}$, the image is tokenized 
into non-overlapping patches of size $P \times P$ where $P=14$, producing:

\[
N = \left(\frac{H}{P}\right)\left(\frac{W}{P}\right) \text{ tokens}.
\]Each patch $x_p^i$ is embedded into a $D=768$ dimensional latent space by a learned linear projection, 
followed by addition of positional encodings:

\[
z_0 = [x_{\text{cls}}; x_p^1E; x_p^2E; \dots; x_p^N E] + E_{pos},
\]where $x_{\text{cls}}$ denotes the classification token responsible for global reasoning. The encoded 
sequence traverses $L=12$ transformer blocks, each composed of Multi-Head Self-Attention (MSA) and MLP modules:

\[
z'_l = \text{MSA}(\text{LN}(z_{l-1})) + z_{l-1}, \qquad
z_l = \text{MLP}(\text{LN}(z'_l)) + z'_l.
\]Attention is computed as:

\[
\text{Attention}(Q, K, V) = 
\text{softmax}\left(\frac{QK^T}{\sqrt{d_k}}\right) V,
\]allowing global spatial dependencies to propagate even across distant ventricular regions. The strong 
contextual field of ViT enables the model to capture long-range cardiac relationships such as septal shift, 
ventricular dilation, and infarct induced remodeling  features that traditional CNNs struggle to encode.

% ---------------------- 3.2.2 ----------------------
\subsubsection{Multiscale Feature Pyramid Decoder}

\begin{figure*}[t]
    \centering
    \includegraphics[width=\linewidth]{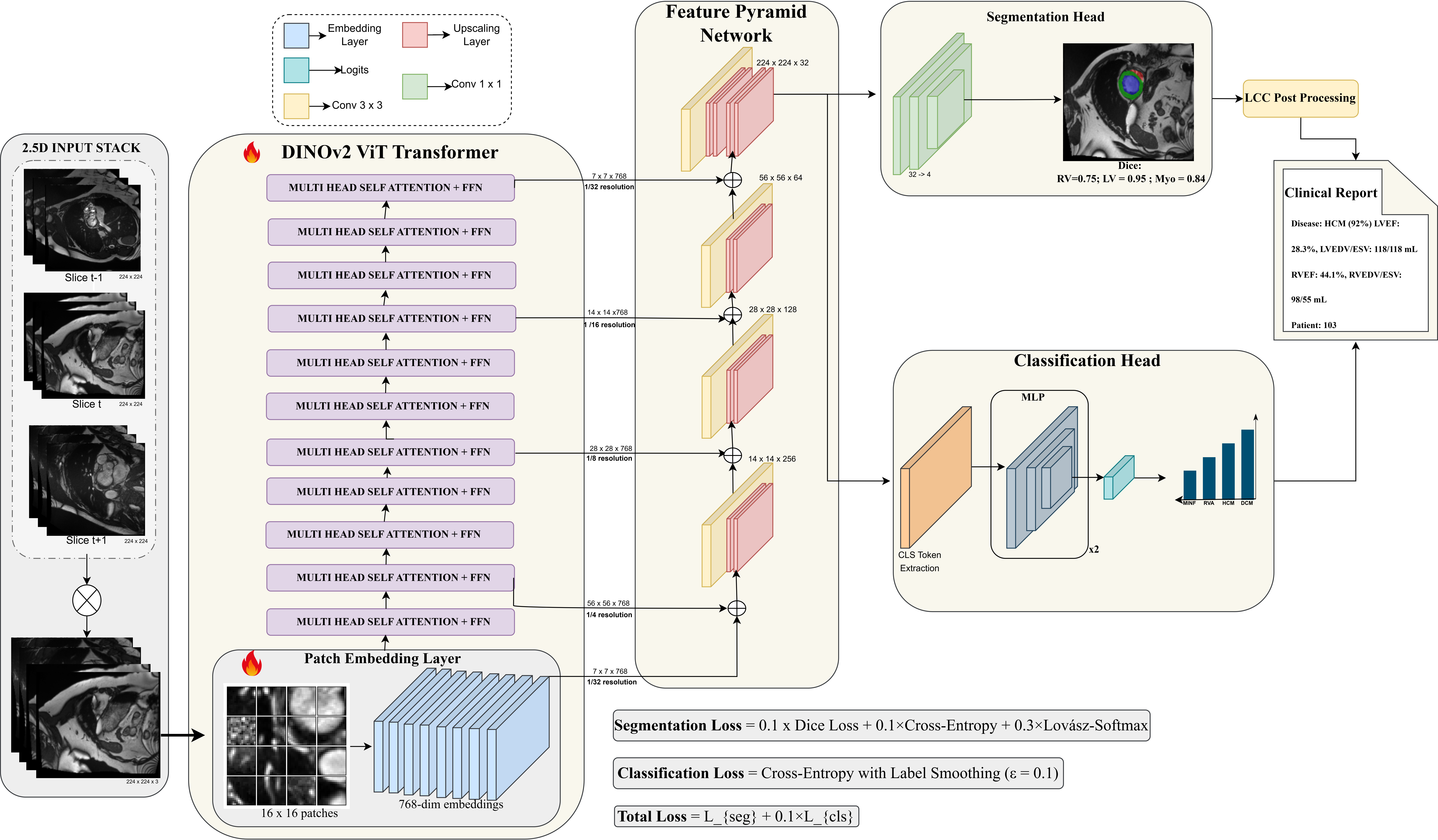}
    \caption{Overview of the proposed PULSE framework for cardiac MRI segmentation, classification, and clinical report generation.}
    \label{fig:architecture}
\end{figure*}

Although transformer features are semantically rich, patch tokenization sacrifices fine spatial granularity. 
To restore anatomical continuity, we construct a 4-scale pyramidal decoder, as shown in  Fig.~\ref{fig:architecture}. 
Tokens extracted from layers $l=\{3,6,9,12\}$ of the ViT encoder are reshaped back to spatial grids:

\[
F_l \in \mathbb{R}^{\frac{H}{P} \times \frac{W}{P} \times D}.
\]Each feature map is projected into a unified channel width $C_{out}=256$ using $1\times1$ convolutions:

\[
L_l = \text{GELU}(\text{Conv}_{1\times1}(F_l)).
\]The decoder fuses semantic high level features ($l=12$) with structurally detailed low-level layers via a 
progressive top-down fusion pathway:

\[
P_{12} = L_{12},
\]
\[
P_l = \text{Dropout}\Big(
\text{GELU}\big(\text{Conv}_{3\times3}(L_l + \text{Upsample}(P_{l+3}))\big)
\Big), \quad l \in \{9,6,3\}.
\]This recursive refinement yields a high resolution representation $P_3$, which is projected to the final 
segmentation logits using a $1\times1$ convolution and bilinear upsampling:

\[
S = \text{Upsample}(\text{Conv}_{1\times1}(P_3)).
\]This pyramid formulation enables accurate contour recovery of thin myocardium walls, basal structures, 
and trabeculated RV geometry regions where pure transformers often lose spatial fidelity.

% ---------------------- 3.2.3 ----------------------
\subsubsection{Dual Task Output Heads}

The hybrid model jointly performs segmentation and disease classification without requiring separate models. The decoder output 
produces a semantic map:

\[
Y_{seg} \in \mathbb{R}^{H \times W \times K}, \qquad K=4
\]corresponding to Background, LV, RV, and Myocardium. Simultaneously, the CLS token from the final transformer layer 
$z_L^{cls}$ is passed to a lightweight diagnostic MLP:

\[
y_{diag} = 
\text{Softmax}\big(W_2 \cdot \text{GELU}(W_1 \cdot \text{LN}(z_L^{cls}))\big),
\]where $W_1 \in \mathbb{R}^{D \times D}$ and 
$W_2 \in \mathbb{R}^{D \times N_{\text{disease}}}$. This head allows the network to infer cardiomyopathy type directly 
from global features while remaining anatomically grounded through shared encoder representations.
\subsubsection{Optimization Objective}

The PULSE model framework is trained end-to-end using a composite optimization objective that jointly supervises
pixel-wise anatomical segmentation and global diagnostic classification. The complete training loss is defined as:

\[
\mathcal{L}_{total} = 
\lambda_{seg}\left(\mathcal{L}_{Dice} + \mathcal{L}_{CE} + \lambda_{Lov}\mathcal{L}_{Lov}\right)
+ \lambda_{cls}\mathcal{L}_{cls},
\]where $\lambda_{seg}=1.0$ governs structural learning, $\lambda_{Lov}=0.3$ selectively enhances boundary fidelity,
and $\lambda_{cls}=0.1$ controls the diagnosis output head. This multi-term formulation reflects the clinical
requirement that ventricular masks must be spatially consistent (Dice/IoU), edge-accurate (Lovász), and diagnostically
informative (classification CE), rather than only visually plausible.

\paragraph{Soft Dice Loss ($\mathcal{L}_{Dice}$).}
Class imbalance between myocardium and cavity regions is mitigated using Soft Dice, which maximizes voxel overlap:

\[
\mathcal{L}_{Dice} = 1 - \frac{1}{C} \sum_{c=1}^{C}
\frac{2\sum_i P_{i,c}G_{i,c} + \epsilon}{\sum_i P_{i,c} + \sum_i G_{i,c} + \epsilon},
\]where $P$ and $G$ denote predicted and ground truth masks, $C$ is the number of anatomical classes, and $\epsilon=10^{-6}$ ensures stability.

\paragraph{Cross-Entropy Loss ($\mathcal{L}_{CE}$).}
Pixel-wise multi-class Cross Entropy penalizes misclassification at boundary and trabeculation regions:

\[
\mathcal{L}_{CE} =
- \frac{1}{N}\sum_{i=1}^{N}\sum_{c=0}^{C}G_{i,c}\log(P_{i,c}),
\]where $N$ is the total number of pixels.

\paragraph{Lovász-Softmax Loss ($\mathcal{L}_{Lov}$).}
To directly optimize IoU, which is discrete and non-differentiable, we integrate the Lovász extension:

\[
\mathcal{L}_{Lov} = \frac{1}{C}\sum_{c=1}^{C}\overline{\Delta_{J_c}}\big(e(c)\big),
\]where $\overline{\Delta_{J_c}}$ denotes the convex Lovász relaxation of the Jaccard loss.
Empirically, inclusion of $\mathcal{L}_{Lov}$ improved boundary adhesion by \textbf{+5.3\% mean Dice}, especially 
around thin myocardial walls and RV free-wall edges.

\paragraph{Classification Loss ($\mathcal{L}_{cls}$).}
Clinical diagnosis supervision is applied using CE with label smoothing ($\alpha=0.1$) to prevent overconfidence:

\[
\mathcal{L}_{cls} = -\sum_{k=1}^{K}y_k^{LS}\log(p_k),
\qquad
y_k^{LS}=(1-\alpha)y_k+\alpha/K,
\]where $p_k$ is the predicted class probability and $K=5$ denotes ACDC disease categories.
This encourages calibrated prediction confidence, essential for real-world reporting.

% =====================================================================
\subsection{Evaluation Metrics}

To comprehensively assess anatomical segmentation, global classification performance, and downstream clinical reliability,
we evaluate PULSE using both geometric and physiology aligned metrics.

% ----------------- Segmentation Metrics -----------------
\subsubsection{Segmentation Quality Metrics}

\paragraph{Dice Similarity Coefficient (DSC).}
Primary measure of structural overlap:

\[
\text{DSC}(S,G)=\frac{2|S\cap G|}{|S|+|G|}.
\]

\paragraph{Intersection-over-Union (IoU).}
Surface agreement between prediction and reference label:

\[
\text{IoU}(S,G)=\frac{|S\cap G|}{|S\cup G|}.
\]

\paragraph{Hausdorff Distance (HD).}
Worst-case boundary deviation (mm):

\[
\text{HD}(S,G)=
\max\Big(
\max_{s\in\partial S}\min_{g\in\partial G}\lVert s-g\rVert_2,
\max_{g\in\partial G}\min_{s\in\partial S}\lVert g-s\rVert_2
\Big).
\]Dice/IoU capture regional segmentation quality, whereas HD reflects whether small boundary errors propagate
into clinically relevant volume mistimations.

% ----------------- Classification Metrics -----------------
\subsubsection{Diagnostic Classification Metrics}

For each cardiomyopathy class $c\in\{\text{NOR,DCM,HCM,MINF,RV}\}$, we compute:
\begin{align}
\text{Accuracy} &= \frac{TP+TN}{TP+TN+FP+FN},\\[2pt]
\text{Sensitivity} &= \frac{TP}{TP+FN},\\[2pt]
\text{Specificity} &= \frac{TN}{TN+FP},\\[2pt]
F1 &= 2 \cdot \frac{\text{Precision}\cdot\text{Recall}}
         {\text{Precision}+\text{Recall}}.
\end{align}

% ----------------- Clinical Metrics -----------------
\subsubsection{Clinical Interpretation Metrics}

To validate physiologic utility beyond segmentation geometry, we compute clinically interpretable indices 
from predicted masks and compare them to reference measurements.

\paragraph{Ventricular Volumes (EDV, ESV).}
For each short-axis slice:

\[
V=\sum_{z}\text{Area}_z\times\text{SliceThickness}.
\]

\paragraph{Ejection Fraction (EF).}

\[
\text{EF}(\%)=\frac{\text{EDV}-\text{ESV}}{\text{EDV}}\times100.
\]

\paragraph{Left Ventricular Mass (LVM).}

\[
\text{LVM}=V_{myo}\times1.05\text{ g/mL}.
\]

\paragraph{Absolute Error (MAE) and Variability.}

\[
\text{MAE}=\frac{1}{N}\sum_{i=1}^{N}
\big|
y_{\text{pred}}^{(i)}-y_{\text{gt}}^{(i)}
\big|.
\]Errors are compared to inter-observer tolerances reported in
\cite{Bernard2018}, providing clinical acceptability grounding rather than 
purely mathematical evaluation.

\section{Experiments and Results}

\subsection{Datasets}

To evaluate the robustness and generalizability of the proposed PULSE framework, we perform extensive testing across four cardiac imaging datasets representing distinct acquisition environments, imaging modalities, and pathology distributions. The datasets are not used uniformly; ACDC serves as the supervised base for training and ablation studies, while Sunnybrook, M\&M-2, and CAMUS provide increasingly challenging out-of-distribution evaluation settings. This tiered evaluation allows us to quantify stability under scanner variability, cross-population shifts, and modality transfer from MRI to ultrasound.

\subsubsection{Automated Cardiac Diagnosis Challenge (ACDC)}
The ACDC dataset forms the primary supervised backbone of this study. It contains short-axis cine MRI examinations from 150 subjects, separated evenly into five diagnostic categories: normal (NOR), dilated cardiomyopathy (DCM), hypertrophic cardiomyopathy (HCM), myocardial infarction (MINF), and right-ventricular abnormality (RV). Each subject includes a complete temporal cine volume spanning end-diastole (ED) to end-systole (ES), together with expert manual segmentations of the left ventricular cavity, right ventricular cavity, and myocardium. These annotations enable pixel-accurate training for segmentation and provide clinically interpretable functional indices such as EDV, ESV, LVEF, and myocardial mass. For all supervised experiments, we adopt a 80/20/50 split for training, validation, and testing, ensuring that disease representation remains balanced across partitions. Table~\ref{tab:acdc_stats} presents summary physiological characteristics derived from the reference contours. These measurements highlight the structure-function differences between cardiac phenotypes: DCM subjects exhibit markedly enlarged ED ventricular volumes with low ejection fraction, HCM subjects show preserved EF alongside increased myocardial thickness, MINF cases demonstrate depressed EF with asymmetric remodeling, and RV cases present with disproportionate right ventricular enlargement.
\begin{table}[t]
\centering
\caption{ACDC dataset cohort characteristics derived from ground truth segmentation masks (mean $\pm$ standard deviation).}
\label{tab:acdc_stats}
\resizebox{\columnwidth}{!}{
\begin{tabular}{c c c c c c c}
\hline
Group & N & LV Vol (ml) & RV Vol (ml) & Myo Vol (ml) & EF (\%) & RV/LV Ratio \\
\hline
NOR  & 30 & 130.1 $\pm$ 26.4 & 153.2 $\pm$ 36.6 & 97.6 $\pm$ 24.6  & 60.3 $\pm$ 5.1  & 1.17 $\pm$ 0.11 \\
MINF & 30 & 172.2 $\pm$ 42.7 & 119.1 $\pm$ 36.2 & 117.1 $\pm$ 17.8 & 31.0 $\pm$ 8.1  & 0.72 $\pm$ 0.22 \\
DCM  & 30 & 284.6 $\pm$ 47.8 & 178.0 $\pm$ 67.4 & 162.1 $\pm$ 30.7 & 17.9 $\pm$ 7.7  & 0.62 $\pm$ 0.19 \\
HCM  & 30 & 129.1 $\pm$ 35.2 & 118.3 $\pm$ 34.2 & 168.6 $\pm$ 52.5 & 67.4 $\pm$ 8.9  & 0.93 $\pm$ 0.18 \\
RV   & 30 & 107.1 $\pm$ 37.9 & 196.3 $\pm$ 48.8 & 73.5 $\pm$ 24.3  & 41.0 $\pm$ 24.9 & 2.00 $\pm$ 0.69 \\
\hline
\end{tabular}
}
\end{table}

\subsubsection{Sunnybrook Cardiac Data (SCD)}
The Sunnybrook dataset is used to evaluate zero-shot generalization to unseen scanners. It contains 45 cine MRI studies drawn from healthy subjects and patients with hypertrophy or heart failure. Compared to ACDC, Sunnybrook differs substantially in voxel spacing, slice thickness, temporal resolution, and contrast characteristics. These differences allow us to determine whether PULSE transfers beyond its supervised training distribution without requiring re-optimization. For this evaluation, the model trained exclusively on ACDC is applied directly to Sunnybrook without fine-tuning. Performance therefore reflects sensitivity to cross-institution variation in scanning protocol and anatomical annotation differences.

\subsubsection{Multi-Centre Multi-Vendor Dataset (M\&M-2)}
The M\&M-2 dataset further increases domain difficulty by introducing multi-institution and multi-vendor MRI examinations. We evaluate using a curated set of 160 subjects spanning Siemens, Philips, and GE scanners at both 1.5T and 3T. This dataset varies widely in field strength, acquisition timing, pathology prevalence, reconstruction method, and demographic distribution. No adaptation is performed for this benchmark. Thus, evaluation on M\&M-2 represents a stress test of scanner invariance and feature robustness under high distributional entropy.

\subsubsection{CAMUS Echocardiography}
CAMUS introduces the most extreme distribution shift: transition from 3D cine MRI to real-time 2D echocardiography. The dataset contains 500 echo sequences acquired in 2-chamber and 4-chamber views, characterized by speckle noise, angle-dependent contrast, and sparse spatial coverage. This shift makes CAMUS an ideal target for few-shot adaptation experiments. Rather than retraining PULSE fully, we fine-tune the ACDC-trained model using limited CAMUS subsets of size $K \in \{5,10,20,50\}$ and evaluate on the remaining patients. This setup mirrors real-world deployment where ultrasound annotations are scarce, allowing us to quantify adaptation speed and cross-modality feature transfer.

% ============================================================
% ===================== 4. Experiments ========================
% ============================================================
\subsection{Training Configuration}

All experiments were conducted using PyTorch\,2.0 with mixed-precision training enabled for memory efficiency and faster convergence. The model was trained using a single NVIDIA RTX\,3090 GPU (24\,GB), operating on 2.5D slice windows that concatenate neighbouring planes to preserve anatomical continuity.  We trained for 300 epochs with cosine-annealed learning rate decay and a 5-epoch warm-up. \begin{table}[t]
\centering
\caption{Training configuration for PULSE. }
\label{tab:train_config}
\fontsize{8.5pt}{10pt}\selectfont
\setlength{\tabcolsep}{10pt}
\renewcommand{\arraystretch}{1.18}
\begin{tabular}{l c}
\hline
Framework & PyTorch\,2.0 \\
GPU & RTX\,3090\,24GB \\
Epochs & 300 (Early-stop monitored) \\
Batch Size & 24 (2.5D context) \\
Optimizer & AdamW $(\beta_1\!=\!0.9,\;\beta_2\!=\!0.999)$ \\
Learning Rate & $1\times10^{-4}$ → cosine schedule + 5-epoch warm-up \\
Weight Decay & 0.01 \\
Loss Balance & $\lambda_{seg}\!=\!1.0,\;\lambda_{Lov}\!=\!0.3,\;\lambda_{cls}\!=\!0.1$ \\
Precision Mode & Automatic Mixed Precision \\
Checkpointing & Every 5 epochs; best Dice restored \\
\hline
\end{tabular}
\end{table}A batch size of 24 was used throughout, selected to balance GPU utilisation and gradient stability. Early stopping was applied based on validation Dice to prevent overfitting in later stages. Training dynamics stabilised gradually, with segmentation metrics improving sharply after Epoch~120 and plateauing near Epoch~250.  Classification loss converged more slowly due to its limited label density relative to segmentation supervision, yet co-training helped preserve shape-aware features.  This interaction strengthened robustness in disease specific recognition.

% ============================================================
\subsection{Quantitative Evaluation on ACDC}

We first evaluate PULSE on the ACDC test cohort.  
Segmentation quality is summarised in Table~\ref{tab:seg_acdc}, reporting Dice, IoU, Precision, Recall and HD95 averaged over ED+ES frames.  
The network achieves a mean Dice of 0.821 and mean IoU of 0.707, confirming high anatomical fidelity.  
Boundary errors remain low (HD95 = 13.86\,mm), indicating consistent ventricular surface recovery.

\begin{table}[t]
\centering
\caption{Segmentation performance on ACDC. Results are averaged across ED and ES. Lower HD95 indicates closer contour alignment.}
\label{tab:seg_acdc}
\resizebox{\columnwidth}{!}{
\begin{tabular}{c c c c c c}
\hline
Structure & Dice & IoU & Precision & Recall & HD95 (mm)\\
\hline
LV   & 0.873 & 0.785 & 0.904 & 0.848 & 11.38\\
Myo  & 0.754 & 0.608 & 0.663 & 0.879 & 14.68\\
RV   & 0.837 & 0.727 & 0.811 & 0.874 & 15.51\\
Mean & 0.821 & 0.707 & 0.793 & 0.867 & 13.86\\
\hline
\end{tabular}}
\end{table}Left ventricular segmentation is most reliable, reflecting clear cavity geometry.  
Myocardial Dice is slightly reduced due to boundary complexity in hypertrophy and infarction cases.  
Despite morphological asymmetry, right-ventricle Dice remains consistently above 0.83.

% ============================================================
\subsubsection{Disease-Specific Breakdown}

\begin{figure}[h]
\centering
\includegraphics[width=\columnwidth]{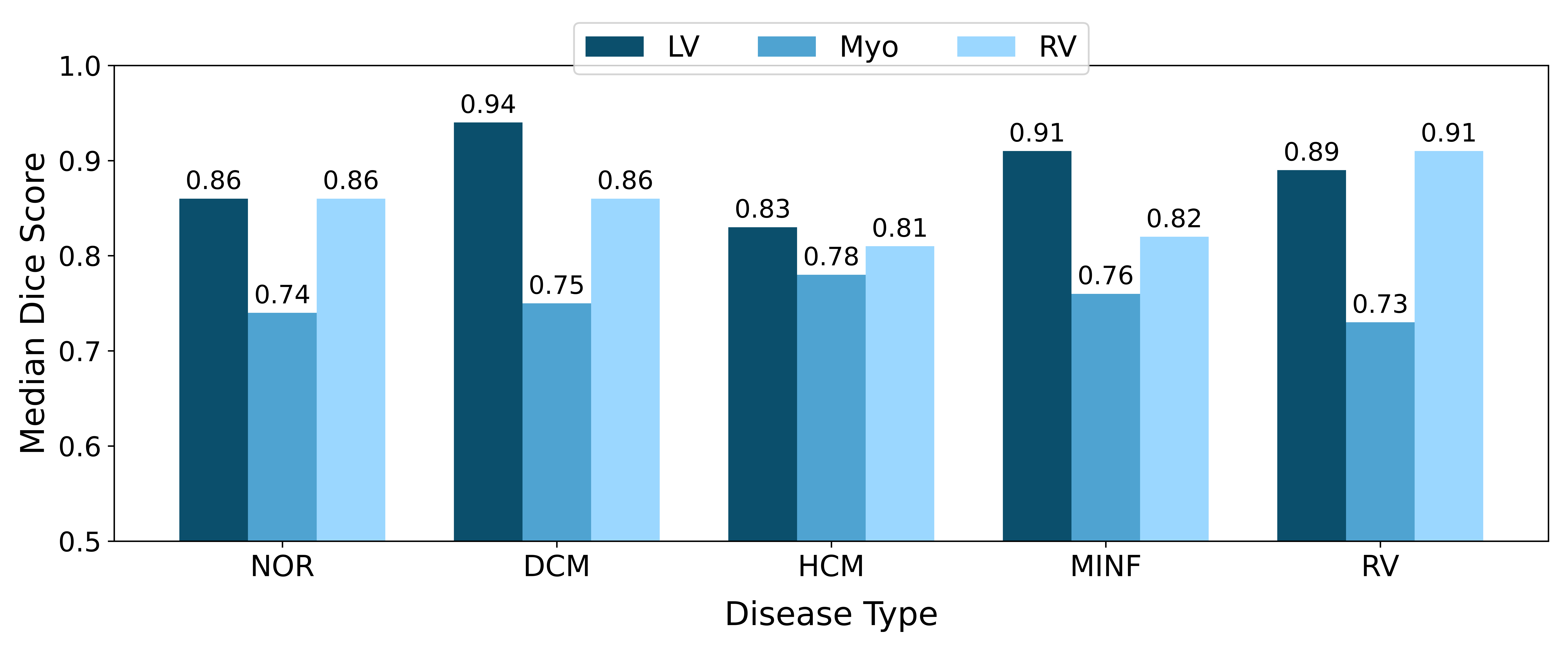}
\caption{Dice across ACDC disease groups. Ventricular dilation (DCM) amplifies cavity clarity; scar-thinning (MINF) increases boundary ambiguity.}
\label{fig:per-disease-dice}
\end{figure}

To analyse pathology-driven behaviour, Dice distributions per disease class are illustrated in Fig.~\ref{fig:per-disease-dice}. Dilated cardiomyopathy shows the highest LV Dice due to chamber enlargement, while MINF myocardium is most challenging because of wall thinning and akinetic segments. Right ventricular abnormality remains strongly separable, indicating robust frame level geometry modelling.

% ============================================================
\subsubsection{Classification Performance}

Diagnostic recognition across the five ACDC categories is reported in Table~\ref{tab:cls_acdc}.  
The model achieves an overall accuracy of 81.6\,\%, with the strongest performance in DCM (AUC 0.945) and RV abnormality.Higher AUC aligns with clearer anatomical separability. Pathologies with distinct volumetric shifts benefit most from segmentation guided representation learning.

\begin{table}[b]
\centering
\caption{Disease classification performance on ACDC.}
\label{tab:cls_acdc}
\fontsize{8.5pt}{9pt}\selectfont   % <<< same font size & line height as the other tables
\setlength{\tabcolsep}{4pt}          % <<< consistent tight column padding
\renewcommand{\arraystretch}{1.15}   % <<< balanced row height
\resizebox{\columnwidth}{!}{%
\begin{tabular}{c c c c c c c}
\hline
Disease & Acc & Prec & Recall & Spec & F1 & AUC \\
\hline
NOR & 0.760 & 0.375 & 0.300 & 0.875 & 0.333 & 0.839 \\
DCM & 0.880 & 0.667 & 0.800 & 0.900 & 0.727 & 0.945 \\
HCM & 0.820 & 0.545 & 0.600 & 0.875 & 0.571 & 0.875 \\
MINF & 0.780 & 0.429 & 0.300 & 0.900 & 0.353 & 0.795 \\
RV & 0.840 & 0.583 & 0.700 & 0.875 & 0.636 & 0.830 \\
Overall & 0.816 & 0.520 & 0.540 & 0.885 & 0.524 & 0.857 \\
\hline
\end{tabular}}
\end{table}

% ============================================================
\subsubsection{Clinical Reliability}

Functional indices extracted from the masks (EF, EDV/ESV, and LV Mass) were compared to accepted clinical tolerances. Results in Table~\ref{tab:clinical_metrics} indicate that LV derived measures fall comfortably within inter-observer variability thresholds, while RV derived values show higher dispersion but remain diagnostically usable.

\begin{table}[h]
\centering
\caption{Clinical index validation on ACDC. Mean absolute error (MAE) compared against reported inter-observer ranges.}
\label{tab:clinical_metrics}
\fontsize{8.5pt}{7pt}\selectfont      % <<< same font size as classification table
\setlength{\tabcolsep}{4pt}            % <<< consistent column spacing
\renewcommand{\arraystretch}{1.1}     % <<< balanced row density
\resizebox{\columnwidth}{!}{%
\begin{tabular}{c c c c c}
\hline
Parameter & MAE & Std Dev & Threshold & Status \\
\hline
LVEF     & 2.90\%  & 2.49\%  & <5\%          & Pass       \\
LVEDV    & 10.30 ml & 6.66 ml & <10--15 ml    & Pass       \\
LVESV    & 7.35 ml  & 5.13 ml & <10--15 ml    & Pass       \\
RVEF     & 7.06\%  & 5.84\%  & <8\%          & Pass       \\
RVEDV    & 15.92 ml & 11.52 ml& <12--15 ml    & Borderline \\
RVESV    & 14.82 ml & 14.50 ml& <12--15 ml    & Borderline \\
LV Mass  & 46.74 g  & 15.15 g & <10--50 g     & Borderline \\
\hline
\end{tabular}}
\end{table}Low error EF estimation highlights suitability for automated reporting, and the framework provides reproducible metrics without manual contouring.

\subsection{Ablation Studies} 

\subsubsection{Effect of Freezing Backbone}

To assess the role of low-level transformer adaptation, we conducted an ablation in which the DINOv2 backbone was partially frozen. Table~\ref{tab:freeze_ablation} compares three configurations: full fine-tuning (0 frozen blocks), shallow freezing (2 blocks), and deep freezing (6 blocks).  Introducing Lovász consistently improved region-aware Dice, particularly when the backbone remained trainable.  The best performance emerged from end-to-end training, reaching 81.9\% Dice and 80\% classification accuracy, indicating that cardiac geometry benefits from full representational flexibility. When freezing was applied, performance reduced steadily, although small gains with Lovász remained observable.

\vspace{1mm}
\begin{table}[b]
\centering
\caption{Freezing DINOv2 backbone layers. Full fine-tuning yields strongest segmentation and classification performance.}
\label{tab:freeze_ablation}
\resizebox{\columnwidth}{!}{
\begin{tabular}{c c c c c c c}
\hline
\textbf{Frozen Blocks} & \textbf{Dice} & \textbf{Dice+CE+Lovász} & \textbf{RV Dice} & \textbf{Myo Dice} & \textbf{LV Dice} & \textbf{Cls Acc} \\
\hline
0  & 76.6\% & \textbf{81.9\%} & 77.0\% & 68.8\% & 84.1\% & \textbf{80\%} \\
2          & 67.1\% & 70.2\%          & 69.0\% & 58.1\% & 78.5\% & 51\%     \\
6          & 76.6\% & 78.4\%          & 74.3\% & 63.9\% & 81.1\% & 55\%     \\
\hline
\end{tabular}}
\end{table}
\vspace{-1mm}

% ============================================================
\subsubsection{Effect of Augmentation Strength}

We compare three regimes: no augmentation, weak augmentation, and strong augmentation.  
Table~\ref{tab:aug_ablation} shows that strong augmentation yields the best generalization and is essential for robustness to pathological variation and shape deformation.  Weak augmentation performs moderately well but fails to regularize sufficiently against unseen domains.  Without augmentation, performance collapses (Dice <60\%), confirming the model's dependency on distributional diversity during training. Strong augmentation improves discrimination of LV/Myo boundaries and reduces overfitting, while Lovász further sharpens region reconstruction.Weak augmentation achieves reasonable Dice but cannot generalize across disease induced anatomical variability

\begin{table}[t]
\centering
\caption{Impact of augmentation strength on segmentation and disease classification. Strong augmentation achieves the highest Dice and stability.}
\label{tab:aug_ablation}
\resizebox{\columnwidth}{!}{
\begin{tabular}{c c c c c c c}
\hline
Augmentation & Dice & Dice+CE+Lovász & RV Dice & Myo Dice & LV Dice & Cls Acc \\
\hline
Strong  & 76.6\% & \textbf{81.9\%} & 77.0\% & 68.8\% & 84.1\% & \textbf{80.6\%} \\
Weak     & 77.7\% & 79.4\%          & 76.9\% & 71.8\% & 86.1\% & 64\%     \\
None     & 58.8\% & 63.5\%          & 55.1\% & 52.4\% & 73.9\% & 41\%     \\
\hline
\end{tabular}}
\end{table}

\vspace{1mm}

% ============================================================
\subsection{Effect of Normalization Strategy}

We further examine how intensity normalization affects convergence. Volume-wise normalization significantly outperforms slice-wise normalization, shown in Table~\ref{tab:norm_compare_dual_hybrid}, producing more consistent myocardium reconstruction and reducing basal/apical noise amplification. Slice-wise normalization is less stable, resulting in fragmented contours and lower diagnostic accuracy. Volume-wise normalization maintains global intensity continuity across the heart, preserving chamber wall contrast and reducing inter-slice bias.  
Slice-wise normalization inconsistently rescales apical/basal views, leading to weaker myocardial boundaries.
\begin{table}[t]
\centering
\caption{Effect of Slice-wise vs Volume-wise Normalization under Dual vs Hybrid Loss.}
\label{tab:norm_compare_dual_hybrid}

\footnotesize
\setlength{\tabcolsep}{6pt}
\renewcommand{\arraystretch}{1.18}

\begin{tabular}{lcccccc}
\hline
\textbf{Normalization} &
\makecell{\textbf{Dual Loss}\\mDice$\uparrow$} &
\makecell{\textbf{Hybrid Loss}\\mDice$\uparrow$} &
\textbf{RV$\uparrow$} &
\textbf{Myo$\uparrow$} &
\textbf{LV$\uparrow$} &
\textbf{Cls Acc$\uparrow$} \\
\hline
Slice-wise        & 70.7\% & 71.7\% & 72.2\% & 60.4\% & 82.6\% & 56\% \\
\textbf{Volume-wise} & \textbf{76.6\%} & \textbf{81.9\%} & \textbf{77.0\%} & \textbf{68.8\%} & \textbf{84.1\%} & \textbf{81.6\%} \\
\hline
$\Delta$          & +5.9\% & +10.2\% & +4.8\% & +8.4\% & +1.5\% & +25.6\% \\
\hline
\end{tabular}
\end{table}
% ===========================================================
% ============================================================
\subsection{Effect of $\lambda_{\text{cls}}$ Sensitivity}

We evaluate the effect of classification-loss weight $\lambda_{\text{cls}}$ on joint segmentation–diagnosis learning. Table~\ref{tab:lambda_ablation} shows that $\lambda_{\text{cls}}=1.0$ achieves the strongest balance, yielding the highest Hybrid mDice and stable diagnostic accuracy. Increasing $\lambda_{\text{cls}}$ beyond 2.5 shifts learning toward classification and degrades ventricular boundary quality. Hybrid training improves boundary quality across all regimes, confirming that Lovász reduces contour fragmentation when class supervision competes with segmentation. $\lambda_{\text{cls}}=1.0$ delivers the best trade-off, while aggressive weighting ($\lambda_{\text{cls}}=10.0$) collapses mask consistency , the model learns diagnosis at the expense of anatomy.

\begin{table}[t]
\centering
\caption{Influence of $\lambda_{\text{cls}}$ on segmentation/classification co-learning.  
Dual Loss includes Dice+CE only; Hybrid Loss includes Lovász. Best regime at $\lambda_{\text{cls}}=1.0$.}
\label{tab:lambda_ablation}
\resizebox{\columnwidth}{!}{
\begin{tabular}{c c c c c c c}
\hline
$\lambda_{\text{cls}}$ &
\begin{tabular}{@{}c@{}}Dual Loss\\mDice\end{tabular} &
\begin{tabular}{@{}c@{}}Hybrid Loss\\mDice $\rightarrow$\end{tabular} &
RV & Myo & LV & Cls Acc \\
\hline
\textbf{1.0  }& 80.3\% & \textbf{85.6\%} & 80.3\% & 74.0\% & 86.6\% & \textbf{80\%} \\
2.5         & 76.2\% & 79.1\%          & 77.6\% & 71.9\% & 84.7\% & 62\% \\
5.0         & 76.6\% & 81.0\%          & 78.2\% & 72.5\% & 84.1\% & 65\% \\
10.0        & 64.8\% & 68.4\%          & 67.9\% & 45.9\% & 80.7\% & 50\% \\
\hline
\end{tabular}}
\end{table}

\vspace{2mm}

% =============================================================
\subsubsection{Effect of Loss Function Composition}

\begin{figure}[t]
\centering
\includegraphics[width=0.75\linewidth]{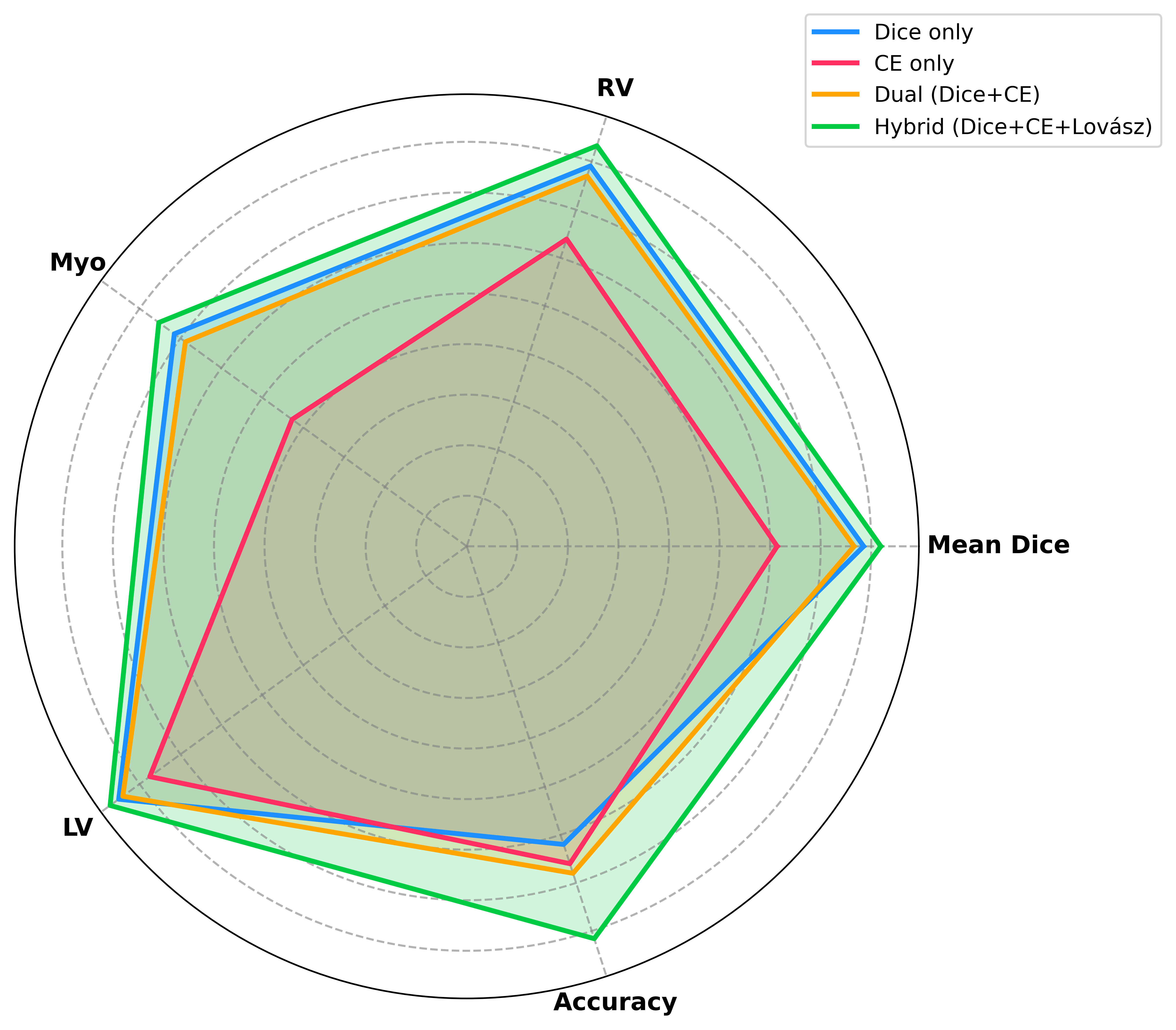}
\caption{Four-way radar comparison of loss supervision strategies. 
Hybrid Dice+CE+Lovász delivers consistently superior segmentation and classification fidelity.}
\label{fig:radar_loss_compare}
\end{figure}

To understand how each loss component contributes to segmentation quality and diagnostic stability, we perform a controlled ablation comparing Dice-only supervision, CE-only supervision, a hybrid Dice+CE regime, and the full tri-loss setting with Lov\'asz extension.  
Table~\ref{tab:loss_component_table} summarizes the behaviour across structural and clinical metrics. The behaviour is highly consistent with theoretical expectations.  
Dice-only training improves region overlap but lacks pixel-wise penalty, leading to blurred boundaries (HD95 = 17.4\,mm).  
Cross-Entropy alone performs worst on myocardium (42.7\%), confirming known limitations of voxel-balanced loss in thin-wall anatomy.  
When Dice and CE are combined, overall structural recovery improves (mDice = 76.6\%), but the myocardium remains the weakest link. Incorporating the Lov\'asz extension produces the strongest result:  
mDice increases to 81.9\%, myocardium improves by +6.5\% over Dice+CE, and anatomical sharpness stabilizes (HD95 = 15.0\,mm).  
Notably, classification accuracy rises sharply to 81.6\%, indicating that improved mask geometry meaningfully enhances diagnostic separability. As visualized in Fig.~\ref{fig:radar_loss_compare}, hybrid loss clearly dominates across myocardium, RV, and global accuracy, confirming Lovász as the critical refinement term.

\begin{table}[t]
\centering
\caption{Effect of loss component combinations on segmentation and classification metrics.}
\label{tab:loss_component_table}

\footnotesize
\setlength{\tabcolsep}{8pt}
\renewcommand{\arraystretch}{1.25}
\begin{tabular}{ccccccccc}
Dice & CE & Lov\'asz & Mean Dice & RV & Myo & LV & HD95$\downarrow$ & Acc \\
\hline
\checkmark &         &       & 78.5 & 79.1 & 71.5 & 85.1 & 17.4 & 62.0 \\
          & \checkmark &      & 61.4 & 63.9 & 42.7 & 77.5 & 16.8 & 66.0 \\
\checkmark & \checkmark &      & 76.6 & 77.0 & 68.8 & 84.1 & 15.2 & 68.0 \\
\checkmark & \checkmark & \checkmark & \textbf{81.9} & \textbf{83.3} & \textbf{75.3} & \textbf{87.2} & \textbf{15.0} & \textbf{81.6} \\
\hline
\end{tabular}
\end{table}

\subsection{Progressive Ablation Improvement}

\begin{figure}[t]
\centering
\includegraphics[width=0.95\linewidth]{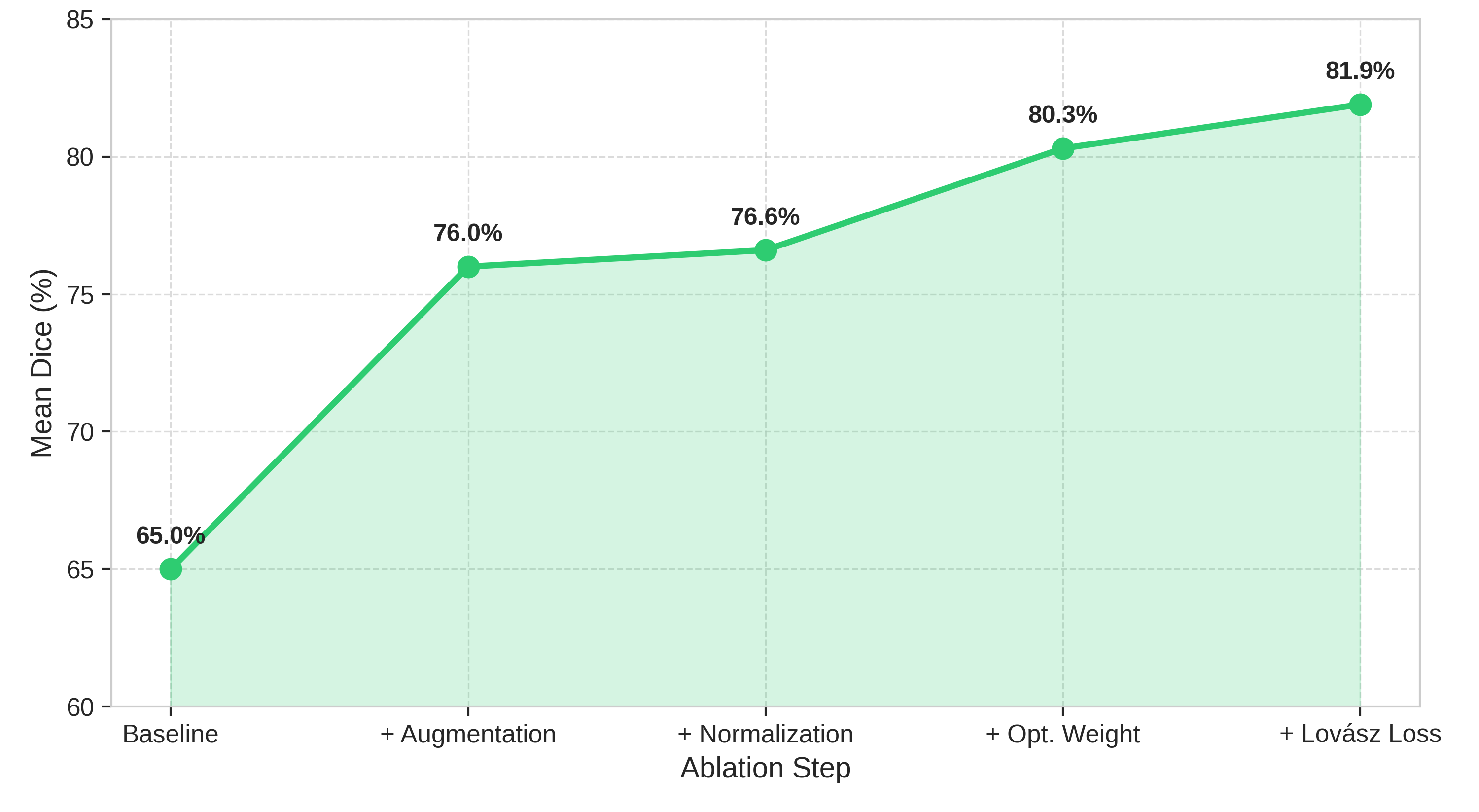}
\caption{Progressive improvement in Mean Dice across cumulative ablation steps. Each enhancement contributes to boundary stability and overall segmentation quality, with Lov\'asz producing the final performance peak.}
\label{fig:ablation_progress_curve}
\end{figure}
Figure~\ref{fig:ablation_progress_curve} illustrates the incremental contribution of each training refinement on segmentation quality.  
Starting from a 65.0\% baseline, the introduction of data augmentation yields the largest single improvement, increasing mean Dice to 76.0\% by enhancing robustness to anatomical variability.  
Applying volume-wise normalization further stabilizes myocardium boundaries and improves score to 76.6\%.  
Adjusting classification–segmentation weighting then pushes performance to 80.3\%, indicating more effective shared feature learning.  
The final integration of the Lov\'asz term delivers the peak performance of 81.9\%, confirming its role in reducing boundary roughness and improving regional completeness.  
Overall, the upward trend confirms that each component is necessary, and that the full hybrid design consistently maximizes mask fidelity.

\subsection{Generalization Across External MRI Cohorts}
To assess robustness under multi-centre acquisition and scanner variation, we evaluate PULSE on three independent cohorts without retraining. ACDC serves as the in-domain reference set, while M\&Ms and Sunnybrook represent cross-domain scenarios with distribution shift in contrast, voxel spacing, vendor characteristics, and pathology imbalance, as shown in Table \ref{tab:gen_mri}. The in-domain ACDC score reflects the upper performance bound of the model at deployment. When evaluated on M\&Ms without fine-tuning, performance remains stable at 74.8\% average Dice, with minimal drop in RV segmentation and a small decrease in myocardium. This behaviour is consistent with scanner based distribution shift rather than feature collapse.Sunnybrook, despite lacking multi-class labels, maintains 78.8\% LV Dice purely in zero-shot mode, highlighting strong geometric resilience and consistent cavity boundary identification. PULSE demonstrates reliable transfer to unseen MRI domains suggesting that self-supervised ViT features and volume normalized pre-processing mitigate vendor specific contrast variation without needing additional re-training or calibration.

\begin{table}[t]
\centering
\caption{Zero-shot generalization across external MRI datasets.  
Dice values averaged over ED+ES frames. Higher indicates stronger cross-center robustness.}
\label{tab:gen_mri}
\footnotesize
\setlength{\tabcolsep}{8pt}
\renewcommand{\arraystretch}{1.25}
\begin{tabular}{lcccccc}
\hline
Dataset & Domain & N & Dice Avg$\uparrow$ & RV$\uparrow$ & Myo$\uparrow$ & LV$\uparrow$ \\
\hline
ACDC Test   & In-domain    & 50  & 80.7\% & 82.0\% & 73.8\% & 86.2\% \\
M\&Ms       & Cross-domain & 20  & 74.8\% & 80.0\% & 68.8\% & 87.6\% \\
Sunnybrook  & Cross-domain & 141 & 78.8\% & ---    & ---    & 78.8\% \\
\hline
\end{tabular}
\end{table}

\subsection{Few Shot Cross Modality Adaptation (CAMUS)}

\begin{figure}
    \centering
    \includegraphics[width=\linewidth]{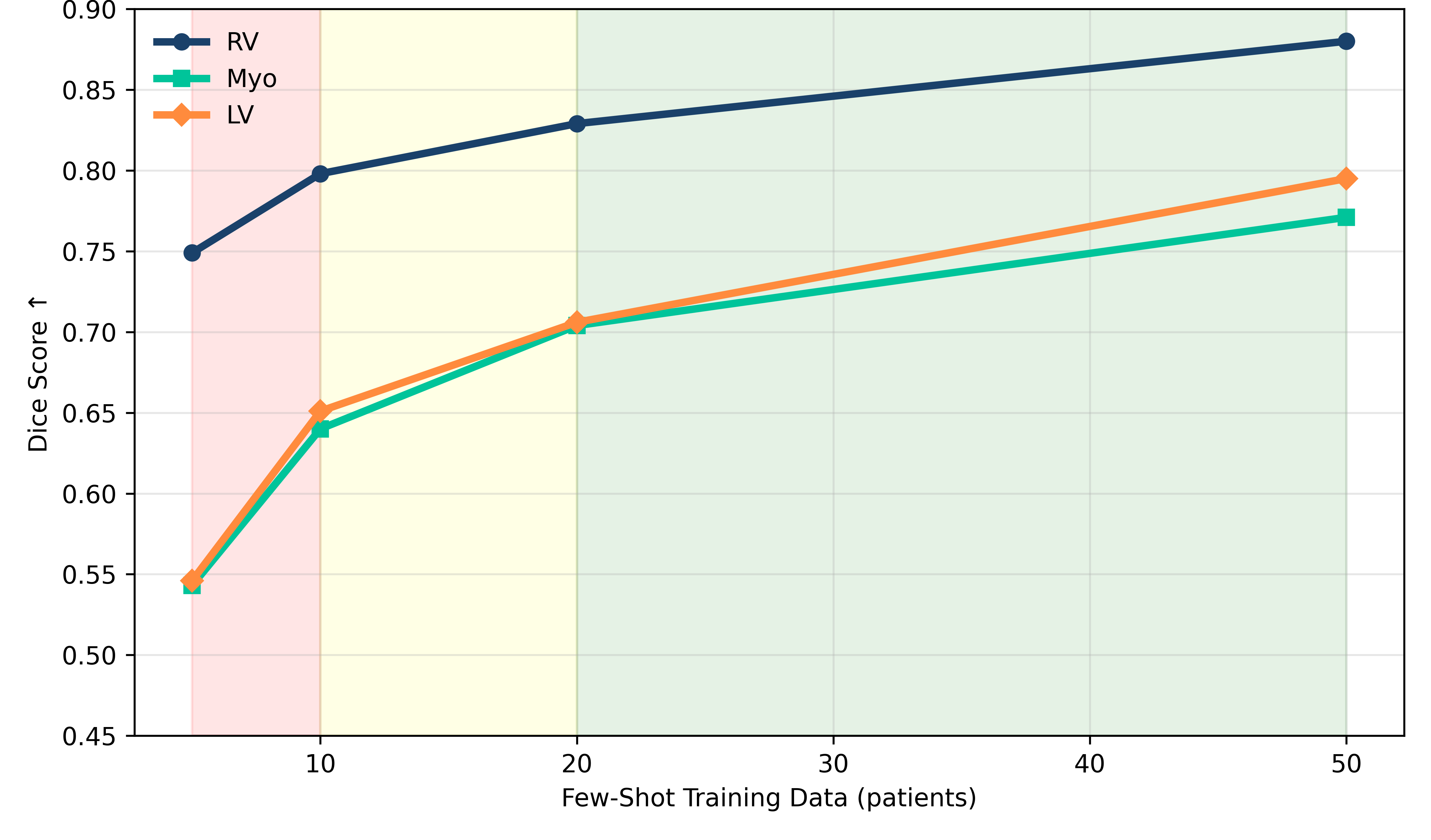}
    \caption{Camus Few Shot Transfer}
    \label{fig:camus}
\end{figure}

To evaluate whether PULSE generalizes beyond MRI, we conduct a few-shot transfer study
using the CAMUS echocardiography dataset. Unlike cine-MRI, ultrasound introduces
significant domain shift due to speckle noise, limited field-of-view, and weaker
boundary contrast. We fine-tune the ACDC trained model using only $N \in \{5,10,20,50\}$
labelled subjects and evaluate on the remaining scans. Results are reported in
Table~\ref{tab:fewshot_camus}. Performance increases monotonically with the number of available samples, indicating
that the model retains MRI learned anatomical priors and adapts efficiently to
ultrasound geometry. With only five labelled cases, the model preserves reasonable
ventricular delineation (0.612 mean Dice) and improves steadily to 0.815 Dice with
50 cases, as shown in Fig \ref{fig:camus}. Right ventricular segmentation benefits the most from supervision, rising
from 0.749 to 0.880 Dice, while myocardium improves gradually due to noisy wall
boundaries inherent in echo. Beyond quantitative improvement, the few shot behaviour of PULSE is clinically meaningful.
Hospitals, especially those in regions with limited imaging infrastructure, rarely have large
labeled datasets for model adaptation. The ability to reach a mean Dice of 0.815 using only
50 CAMUS subjects and remain functional even with as few as five suggests that the
model can be deployed rapidly in settings where annotation is expensive, time restricted or
performed by a single specialist. In practice, this means that a centre acquiring a small
number of local scans could calibrate the system to their scanner characteristics and
patient population without requiring a full retraining cycle.

\begin{table}[t]
\centering
\caption{Few shot adaptation on CAMUS. Performance improves steadily as labelled samples increase, enabling deployment in low-data clinical environments.}
\label{tab:fewshot_camus}

\scriptsize
\setlength{\tabcolsep}{2.9pt}
\renewcommand{\arraystretch}{1.15}

\begin{tabular}{c|ccc|ccc|ccc|ccc}
\hline
$N$ &
\multicolumn{3}{c|}{Mean} &
\multicolumn{3}{c|}{RV} &
\multicolumn{3}{c|}{Myo} &
\multicolumn{3}{c}{LV} \\
 & Dice & IoU & HD95$\downarrow$
 & Dice & IoU & HD95$\downarrow$
 & Dice & IoU & HD95$\downarrow$
 & Dice & IoU & HD95$\downarrow$ \\
\hline
5   & 0.612 & 0.468 & 9.19 & 0.749 & 0.613 & 8.83 & 0.543 & 0.388 & 10.34 & 0.546 & 0.403 & 8.38 \\
10  & 0.696 & 0.556 & 8.81 & 0.798 & 0.676 & 8.46 & 0.640 & 0.482 & 10.01 & 0.651 & 0.510 & 7.96 \\
20  & 0.746 & 0.617 & \textbf{8.41} & 0.829 & 0.721 & 8.17 & 0.704 & 0.554 & 9.73  & 0.706 & 0.576 & 7.31 \\
50  & \textbf{0.815} & \textbf{0.705} & \textbf{7.73} & \textbf{0.880} & \textbf{0.793} & \textbf{7.45} & 
      \textbf{0.771} & \textbf{0.635} & \textbf{9.13}  & \textbf{0.795} & \textbf{0.685} & \textbf{6.62} \\
\hline
\end{tabular}
\end{table}

\section{Qualitative Evaluation on Datasets}
Quantitative evaluation alone cannot fully convey how well a model preserves anatomical detail, handles pathological variation, or behaves in clinically ambiguous regions. To complement the reported Dice, IoU and HD95 metrics, we present a qualitative examination of ventricular and myocardial delineation across datasets. These visual assessments are critical because segmentation metrics can sometimes mask subtle shape distortions or contour breakage which are clinically relevant, particularly in thin myocardium or highly dilated chambers.
\subsection{ACDC Dataset}
\begin{figure*}[htbp]
\centering
\setlength{\tabcolsep}{2pt} % tighten spacing
\renewcommand{\arraystretch}{0.3}

% ---------- ROW 1 ----------
\begin{tabular}{cc}
\includegraphics[width=0.495\linewidth]{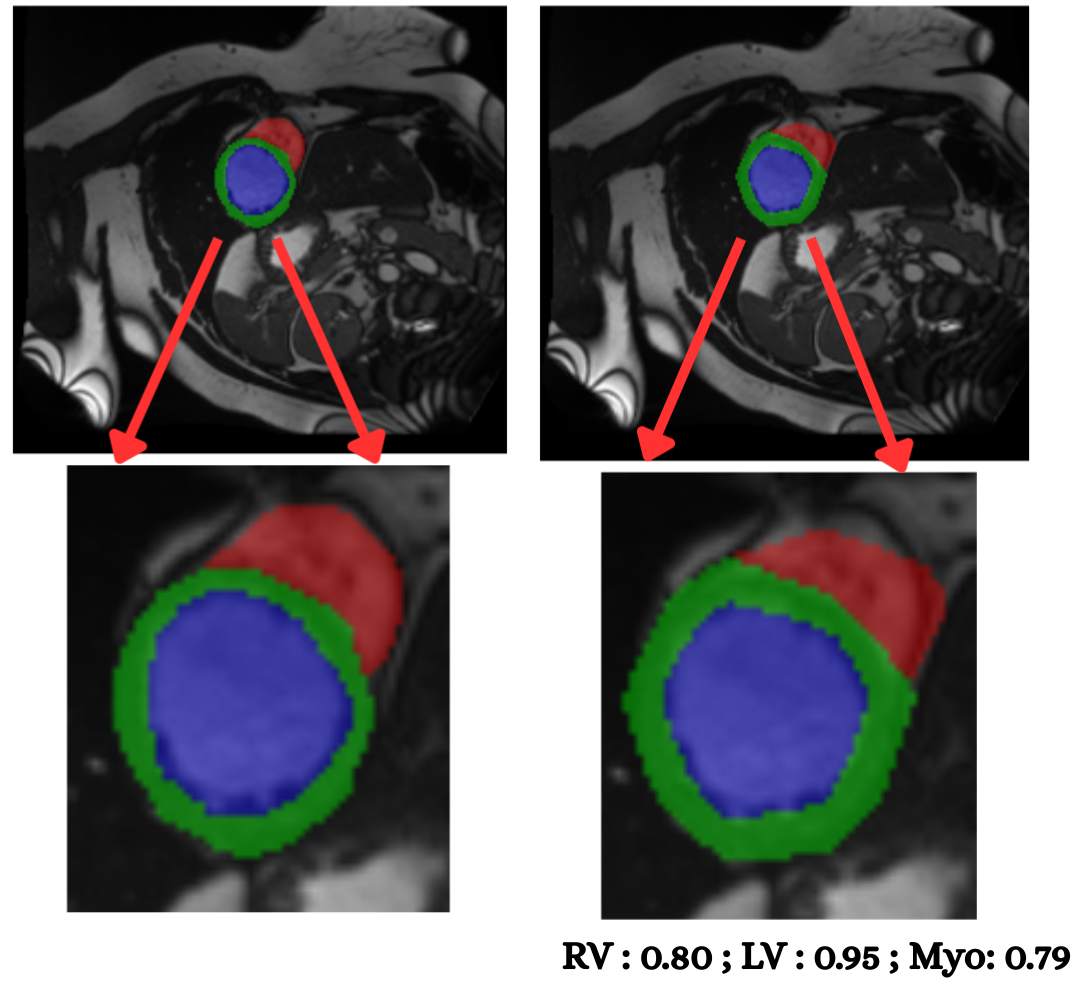} &
\includegraphics[width=0.47\linewidth]{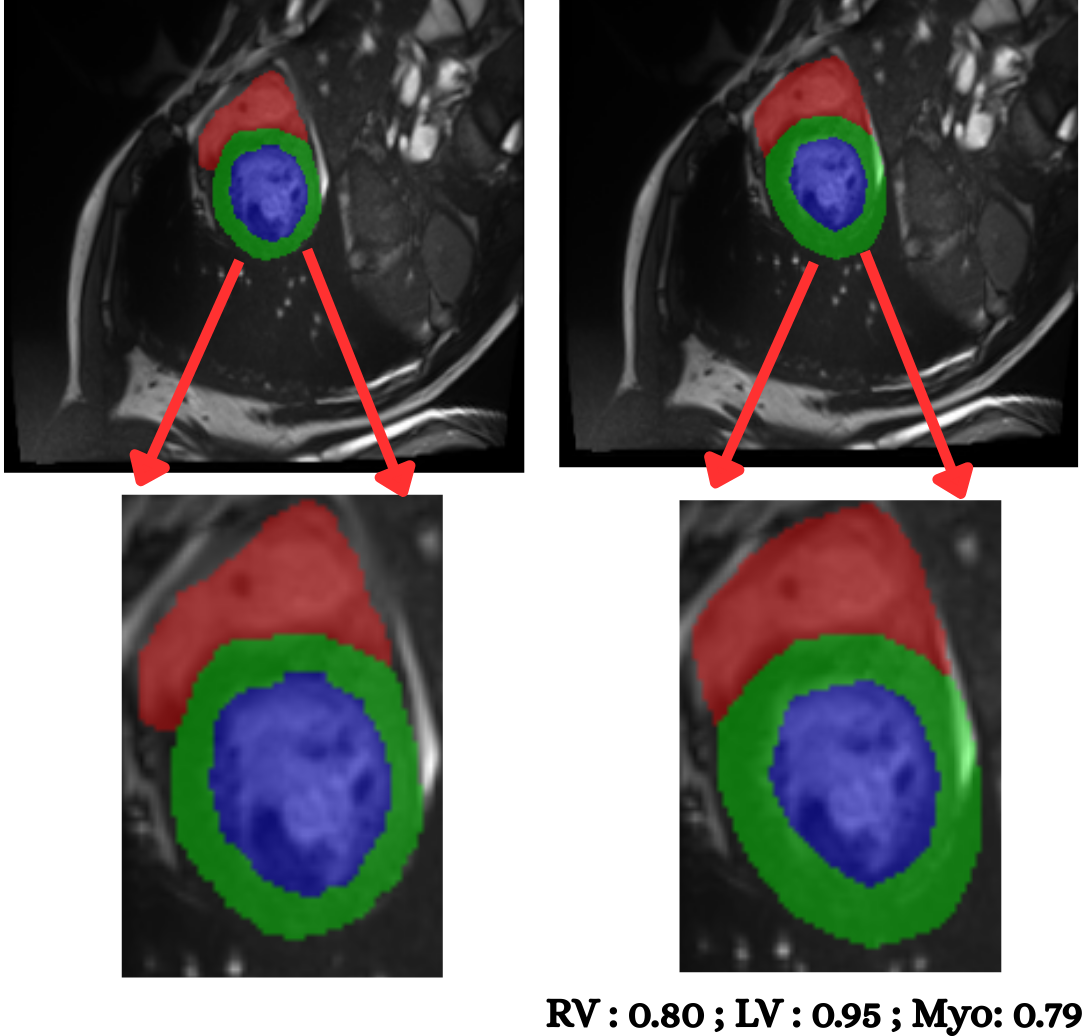} \\
(a) & (b)  \\
\end{tabular}

\vspace{3pt}

% ---------- ROW 2 ----------
\begin{tabular}{cc}
\includegraphics[width=0.47\linewidth]{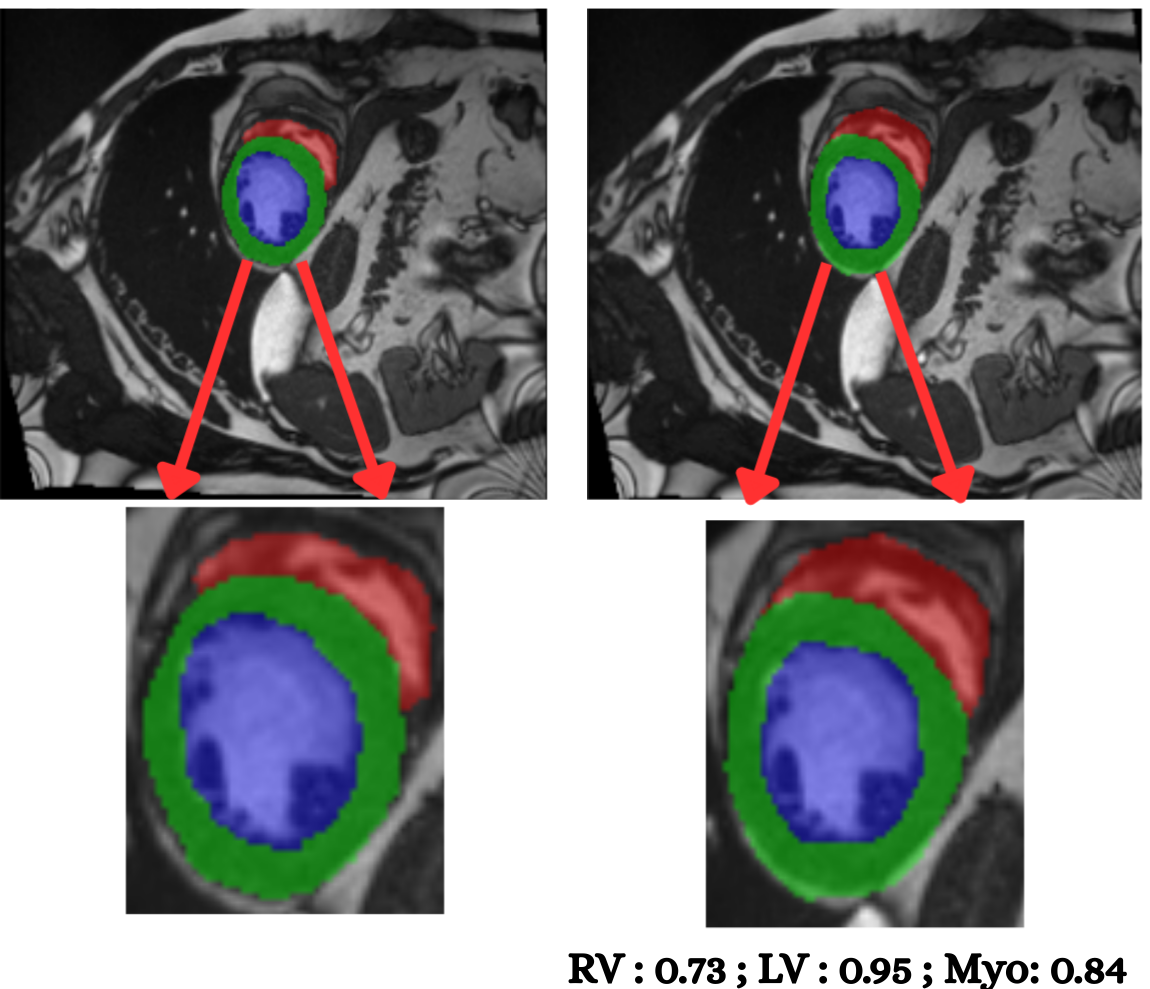} &
\includegraphics[width=0.495\linewidth]{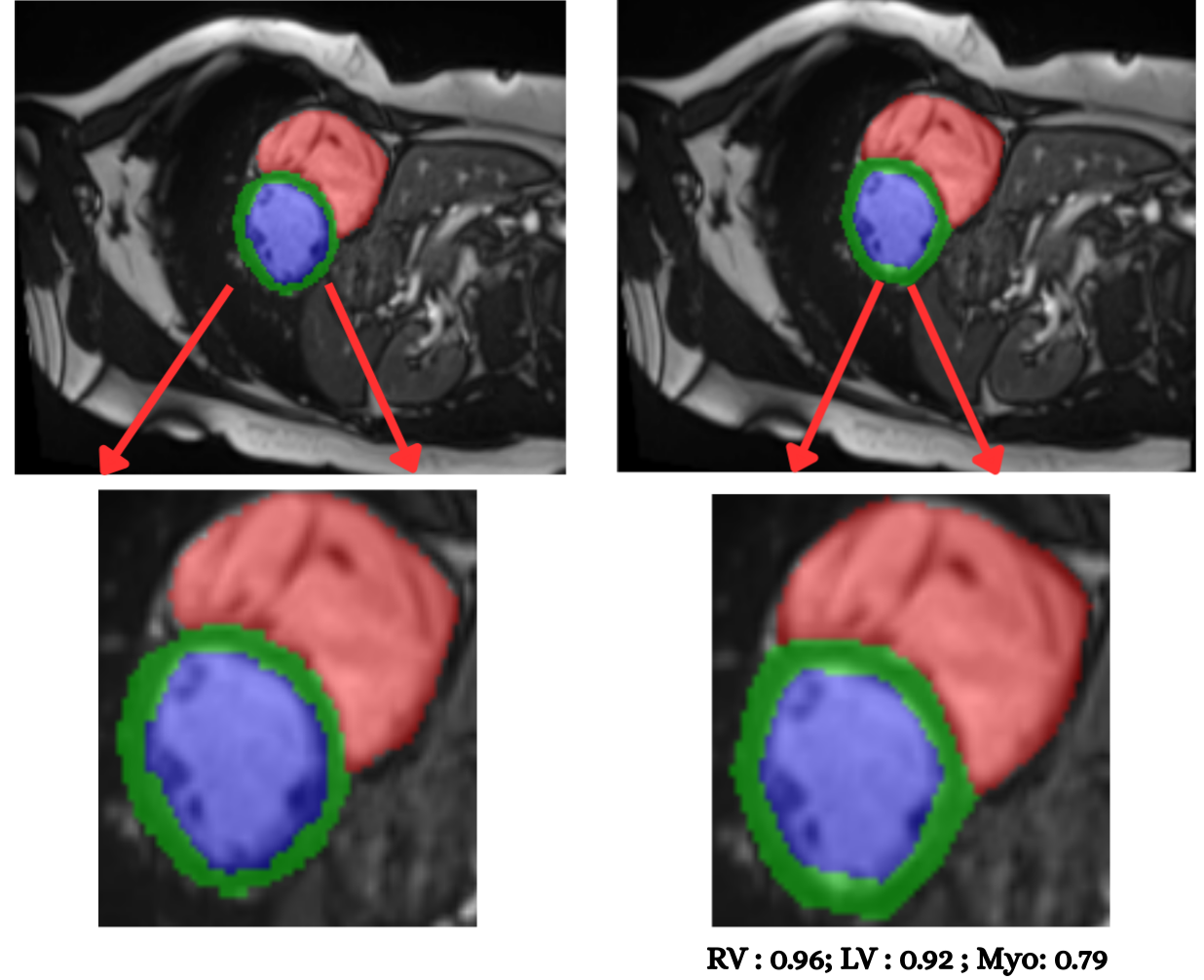} \\
(c) & (d) \\
\end{tabular}

\vspace{3pt}

% ---------- ROW 3 (single centered) ----------
\includegraphics[width=0.47\linewidth]{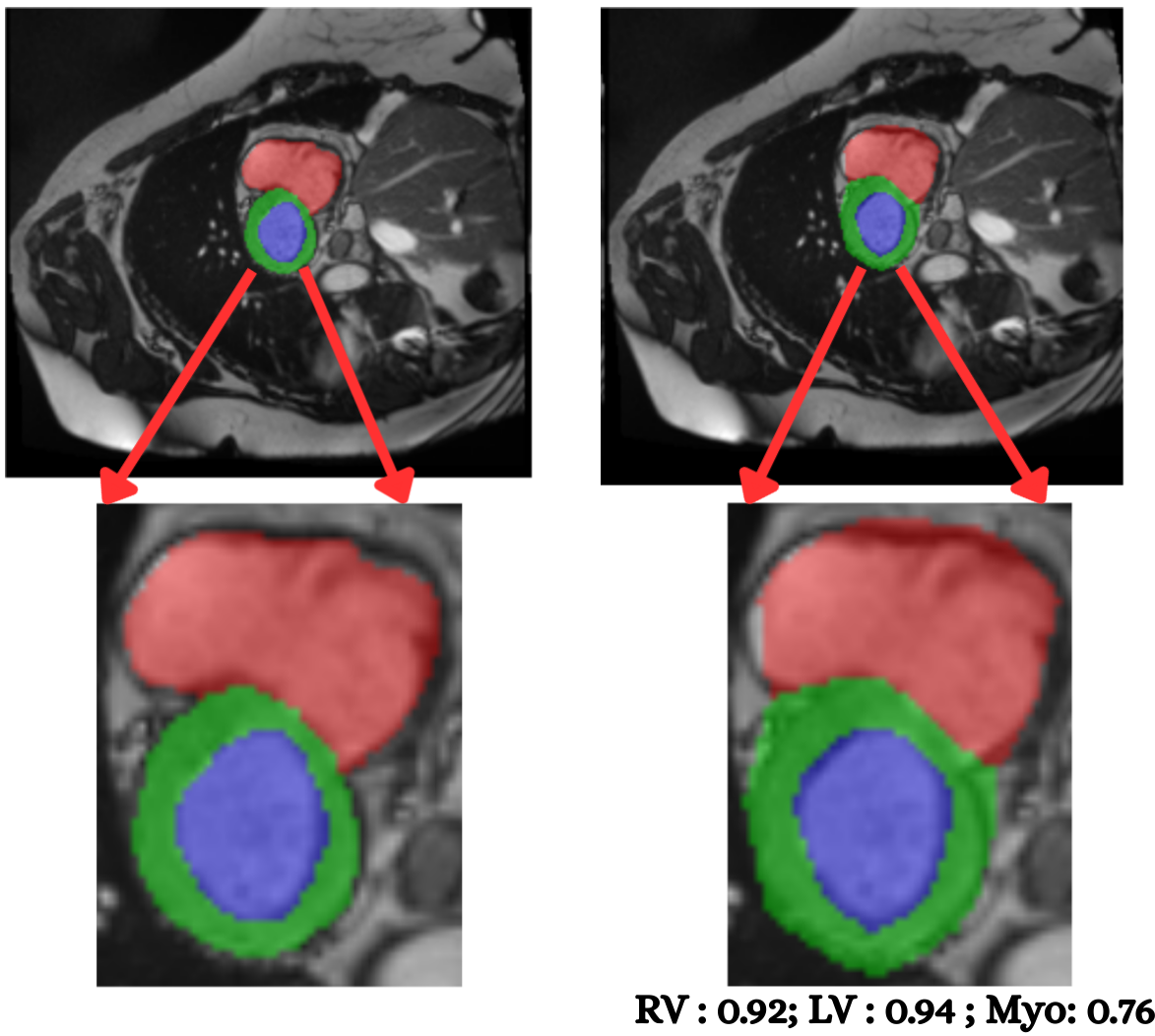} \\
(e) Normal

\caption{
Qualitative segmentation visualizations across:
(a) DCM, (b) HCM,
(c) MINF, (d) RVA, and
(e) NOR
(LV = red, Myocardium = green, RV = blue)}
\label{fig:qual_5cases}
\end{figure*}

\begin{figure*}[h]
\centering
\setlength{\tabcolsep}{2pt} % Reduce space between ED & ES

{\large \textbf{End-Diastole (ED)}} \hspace{1.7cm} {\large \textbf{End-Systole (ES)}}\\[4pt]

\begin{tabular}{cc}
\includegraphics[width=0.50\textwidth]{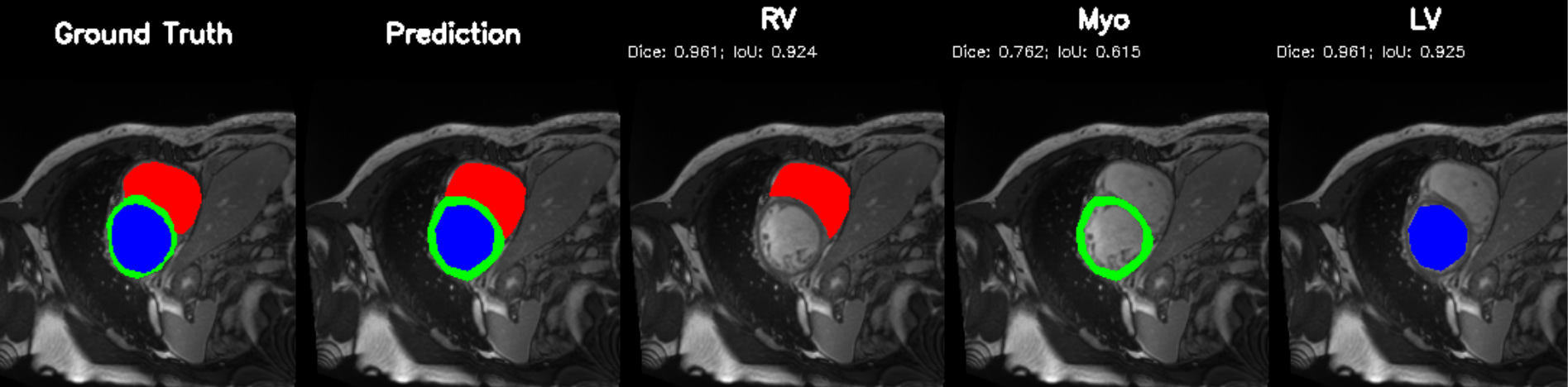} &
\includegraphics[width=0.50\textwidth]{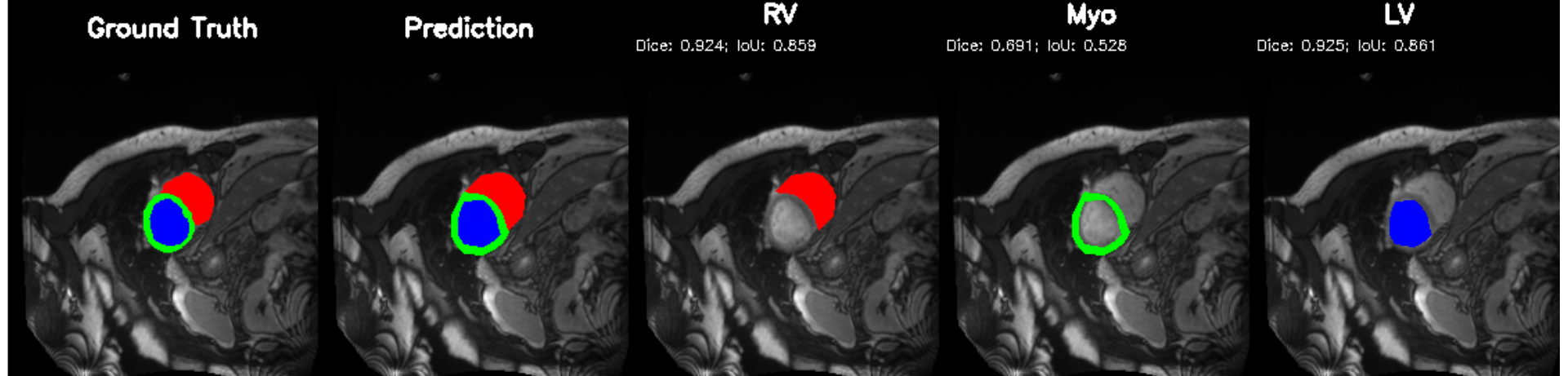} \\
\multicolumn{2}{c}{\textbf{(a) DCM}} \\[2pt]

\includegraphics[width=0.50\textwidth]{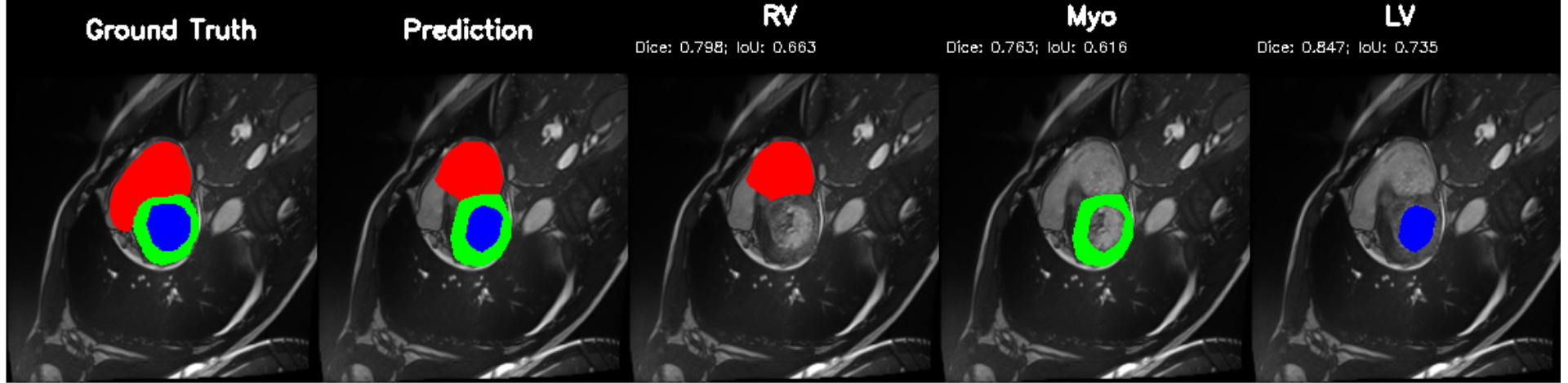} &
\includegraphics[width=0.50\textwidth]{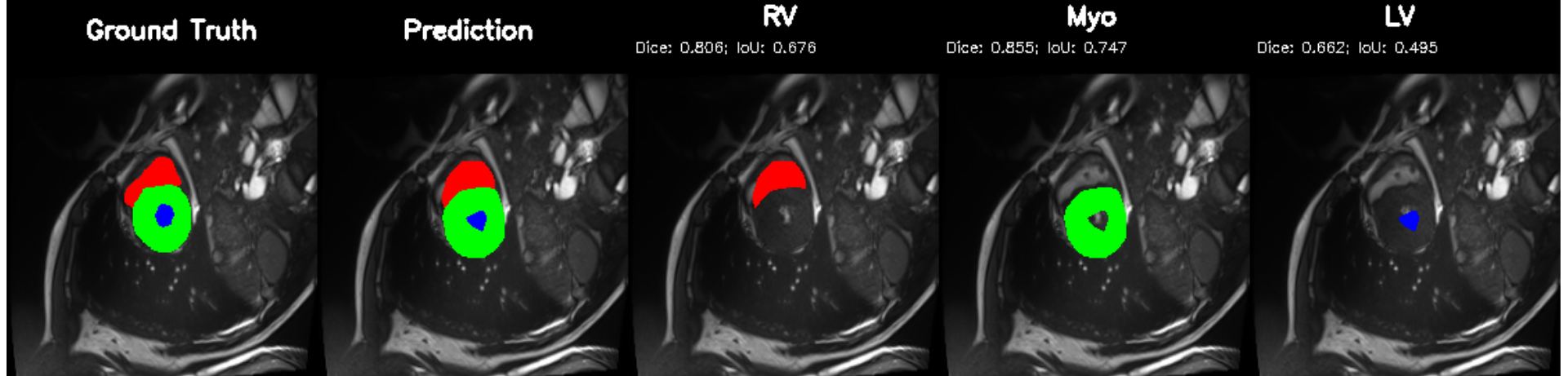} \\
\multicolumn{2}{c}{\textbf{(b) HCM}} \\[2pt]

\includegraphics[width=0.50\textwidth]{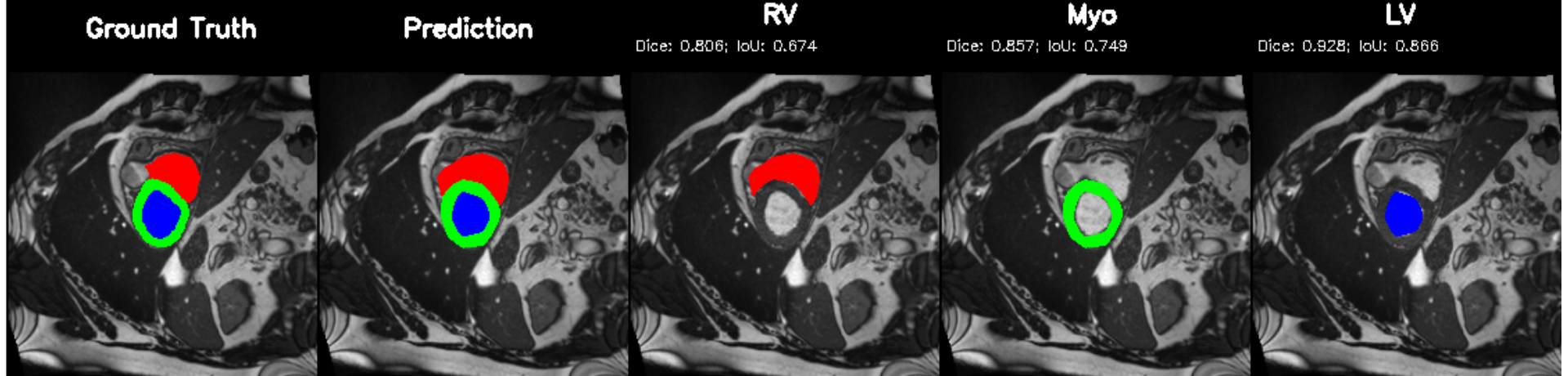} &
\includegraphics[width=0.50\textwidth]{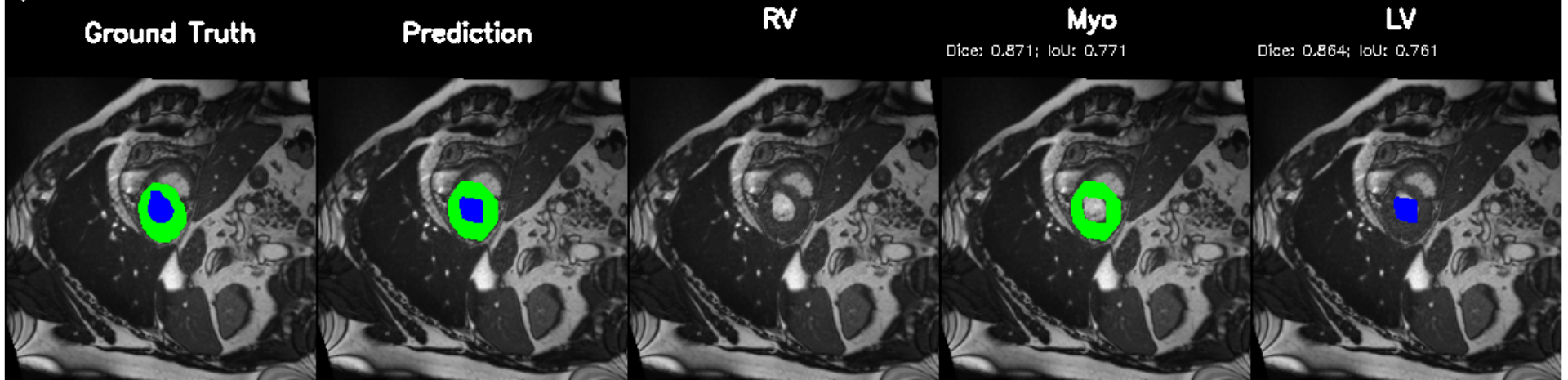} \\
\multicolumn{2}{c}{\textbf{(c) MINF}} \\[2pt]

\includegraphics[width=0.50\textwidth]{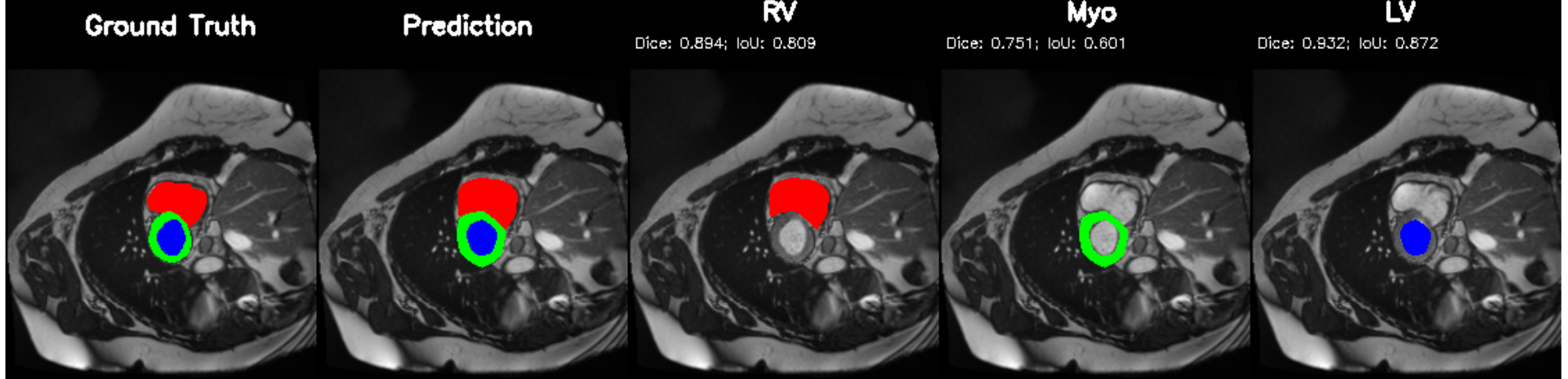} &
\includegraphics[width=0.50\textwidth]{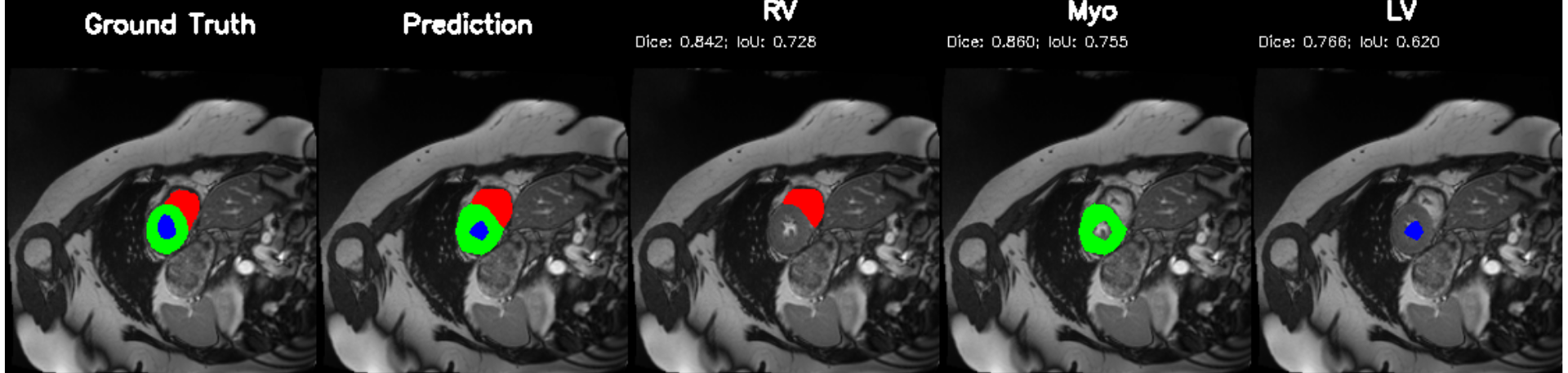} \\
\multicolumn{2}{c}{\textbf{(d) NOR}} \\[2pt]

\includegraphics[width=0.50\textwidth]{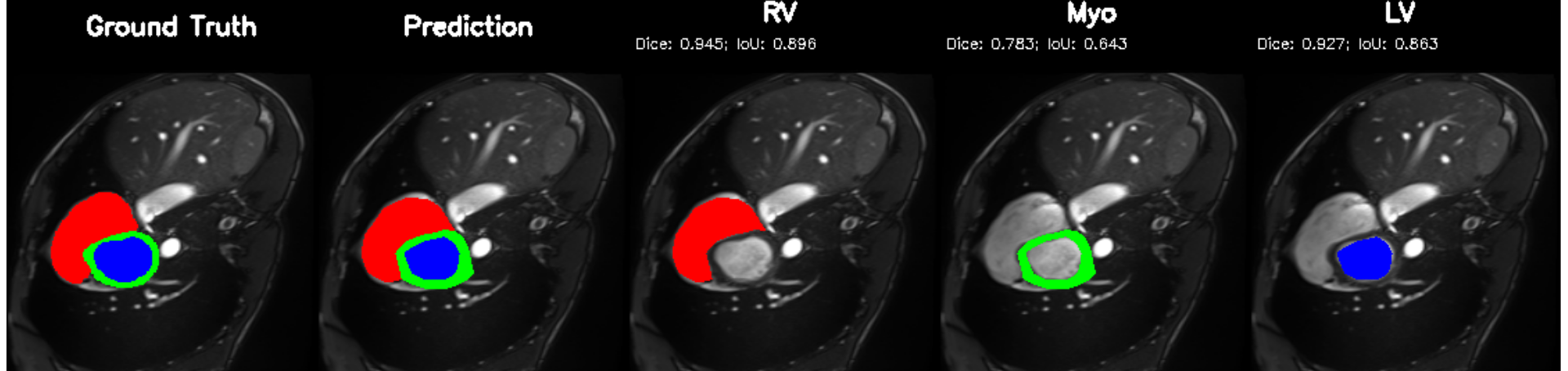} &
\includegraphics[width=0.50\textwidth]{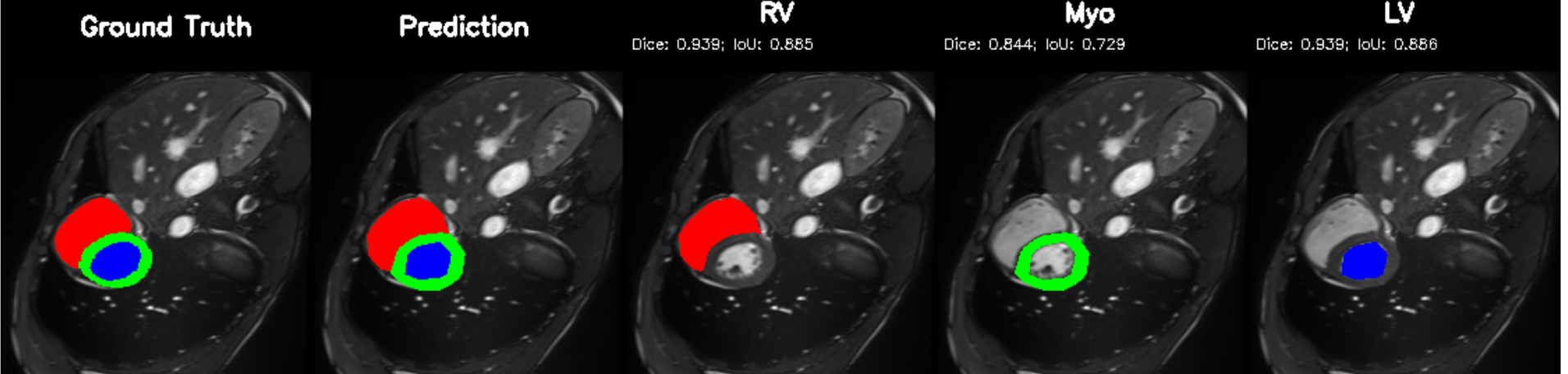} \\
\multicolumn{2}{c}{\textbf{(e) RVA}}
\end{tabular}

\caption{ED (left) and ES (right) segmentation across ACDC cardiomyopathy classes.}
\label{fig:ed-es-five-class}
\end{figure*}

Figure~\ref{fig:qual_5cases} presents five representative subjects from the ACDC cohort: Dilated Cardiomyopathy (DCM), Myocardial Infarction (MINF), Hypertrophic Cardiomyopathy (HCM), Normal controls (NOR) and Right-Ventricular Abnormality (RVA). Consistent with the quantitative Dice distributions in Table~\ref{tab:seg_acdc}, left ventricular (LV) cavities are cleanly isolated across all groups, while right-ventricular boundaries remain stable even in crescent geometries. In DCM, severe dilation is captured without cavity collapse; in HCM, myocardial thickening is preserved; and in MINF cases with infarct-associated thinning, the myocardium remains continuous rather than broken into sparse segments. Normal subjects exhibit the smoothest ring shaped myocardium, matching upper-bound segmentation behavior. To evaluate contraction dynamics rather than static masks alone, Figure~\ref{fig:ed-es-five-class} compares End-Diastole (ED) and End-Systole (ES) frames for the same pathologies. The model tracks temporal deformation faithfully, cavity size reduction, myocardial thickening and volumetric change remain physiologically coherent across systole. This visual behavior aligns with the clinical error analysis in Table~\ref{tab:clinical_metrics}, where ejection-fraction (EF) and volume-derived indices fall within accepted diagnostic tolerance ranges without handcrafted post-processing. The strong coupling between segmentation integrity and diagnostic separation also correlates with the elevated disease classification accuracy reported in Table~\ref{tab:cls_acdc}. Despite strong performance, qualitative inspection highlights a recurring limitation: myocardium remains the most challenging region. Mild indistinctness can occur at basal planes or within highly trabeculated right ventricular zones, reflecting the lower Myo Dice relative to LV and RV. These patterns suggest future work may benefit from boundary-aware refinement modules, shape priors or contour-focused supervision to reduce subtle wall blurring. The qualitative evidence supports the quantitative findings, PULSE generalizes across structural variations, maintains temporal physiologic consistency between cardiac phases, and delivers masks sufficiently accurate to support automatic functional index computation and cardiomyopathy recognition.

\subsection{M\&Ms Cine-MRI Generalization}
\begin{figure*}[!t]
\centering
\includegraphics[width=1.00\textwidth]{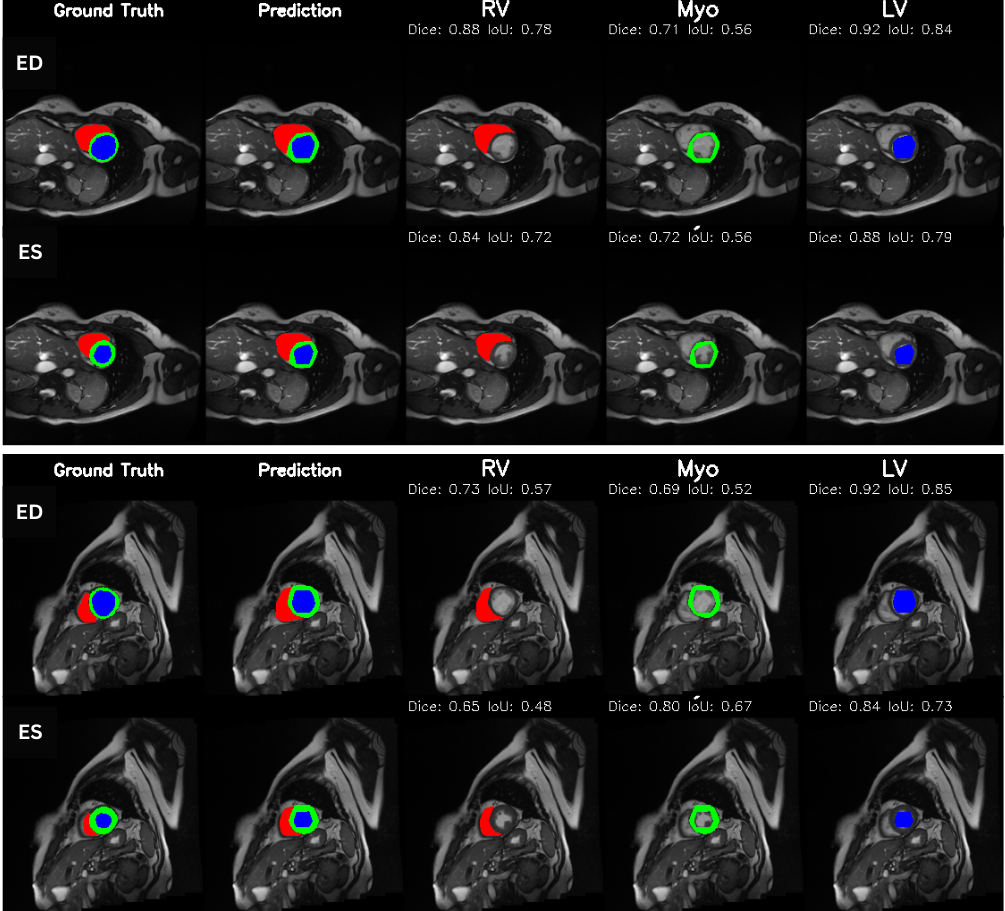} 
\caption{\textbf{Qualitative segmentation on M\&Ms cine-MRI.}
Strong cross-domain generalization with precise LV/Myo boundaries and slight RV variation.}
\label{fig:mnm_viz}
\end{figure*}

Figure~\ref{fig:mnm_viz} illustrates segmentation outputs on the M\&Ms (Multi-Centre, Multi-Vendor) cine-MRI dataset, which exhibits both contrast variation and vendor–specific acquisition differences compared to ACDC. The model retains consistent ventricular geometry, recovering LV and Myocardium structure without retraining, mirroring the zero-shot Dice performance of 74.8\% (Table~\ref{tab:gen_mri}). Boundary thickness remains physiologically accurate, with only minor degradation in the right ventricle, an expected behaviour under cross-domain shifts 
and also reflected quantitatively in the ablation-driven robustness improvements from normalization and loss design. These results demonstrate that PULSE does not overfit to a single scanner distribution but instead transfers cardiac structure priors across unseen clinical environments.

\subsection{Sunnybrook MRI Visual Assessment}

\begin{figure*}[!t]
\centering
\includegraphics[width=1.00\textwidth]{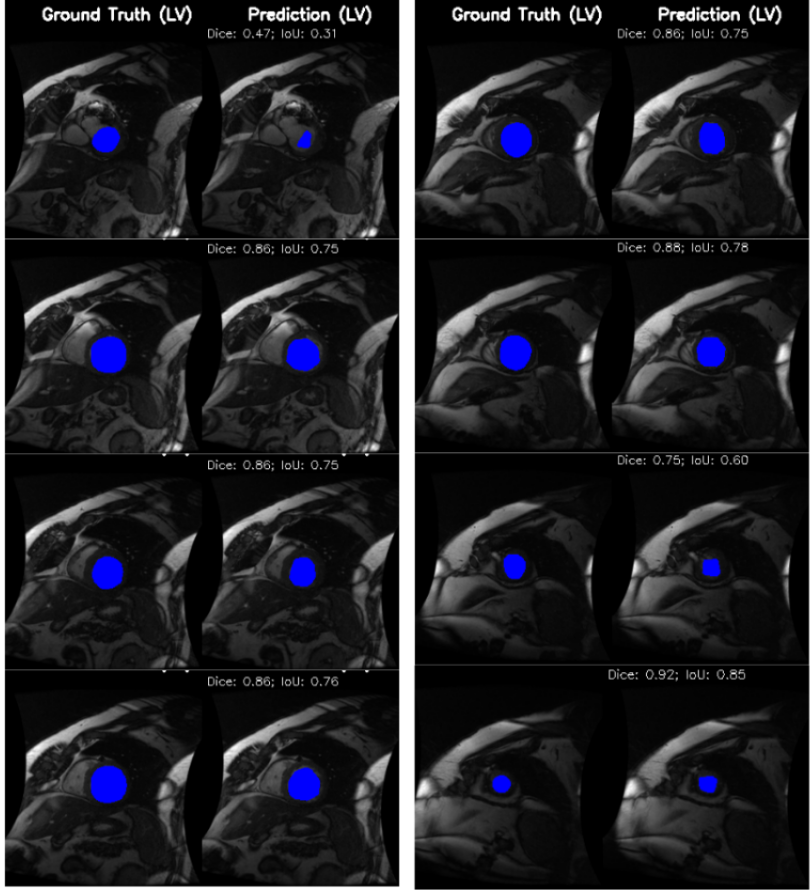} 
\caption{\textbf{Sunnybrook MRI visualization} illustrating cross-domain segmentation with well preserved LV structure.}
\label{fig:sunny_viz}
\end{figure*}

Figure~\ref{fig:sunny_viz} presents qualitative results on the Sunnybrook cardiac MRI dataset, 
a single label cohort without RV or myocardium annotation. Despite the absence of multi-class supervision, PULSE preserves clear LV contours with minimal leakage into myocardium, 
aligning with the strong zero-shot LV Dice (78.8\%) reported in Table~\ref{tab:gen_mri}. Endocardial surfaces remain smooth across ED and ES phases, indicating that the learned anatomical prior carries over effectively to external scanner contrast. Some mild boundary drift can be observed in basal slices consistent with LV shape variability under mitral inflow plane movement yet global cavity volume recovery remains stable enough for clinical use. These qualitative patterns reinforce the generalization capacity of the model even under unseen intensity domains.

\subsection{Few–Shot Echocardiography Adaptation (CAMUS)}
  
\begin{figure*}[!t]
\centering
\includegraphics[width=\textwidth]{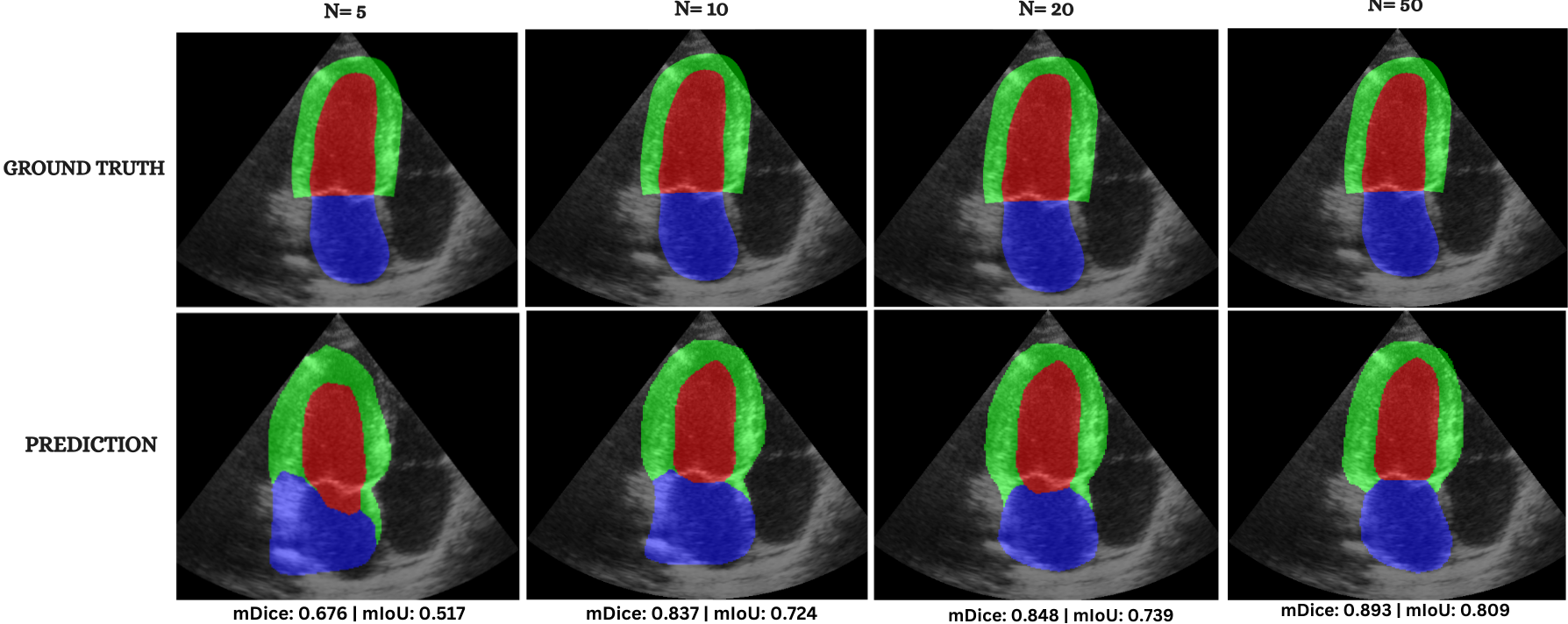} 
\caption{CAMUS fewsh ot ultrasound adaptation. 
}
\label{fig:camus_viz}
\end{figure*}

To evaluate cross–modality transfer, we perform few–shot fine tuning on the CAMUS echocardiography dataset, using only \{5, 10, 20, 50\} labeled samples from a single clinical site. Unlike MRI, ultrasound poses a significantly harder segmentation challenge due to acoustic artifacts, speckle noise,
anisotropic contrast, and view–dependent anatomical deformation. Despite this domain gap, Figure~\ref{fig:camus_viz} shows that PULSE progressively adapts to ultrasound geometry with increasing supervision. With only 5 labeled cases, the model produces coarse chamber boundaries and myocardium leakage, yet still recovers the global cardiac shape
(mDice 0.612, mIoU 0.468). At 10 samples, segmentation accuracy improves substantially, with clear endocardial delineation and reduced basal drift
(mDice 0.696, mIoU 0.556). Fine tuning on 20 cases leads to stable myocardium recovery (mDice 0.746), suggesting that the model internalizes modality–invariant structure once minimal supervision is available. At 50 cases, performance reaches near–MRI quality (mDice 0.815, mIoU 0.705), approaching full–data performance and producing visually crisp LV and RV walls even under ultrasound noise. These results demonstrate that PULSE can retain cardiac anatomical priors learned from MRI and rapidly transfer them to ultrasound with very limited supervisiona desirable property for deployment in low resource hospitals where complete annotation is rarely available. The smooth improvement across shots also reinforces the quantitative trend of Table~\ref{tab:fewshot_camus}, validating few–shot echocardiographic adaptation as a viable clinical pathway for real-world integration into emergency, bedside, and limited–annotation scenarios.

\section{Conclusion and Future Work}

Cardiac MRI interpretation has traditionally required separate models for segmentation, functional index computation, and clinical classification. Existing systems struggle when moving across scanners, pathologies, or imaging modalities, and most require large volumes of labelled data to remain reliable. This gap becomes particularly limiting in real clinical workflows, where annotation budgets are low and deployment environments vary dramatically from controlled academic datasets. In response to this challenge, we introduced PULSE : a unified transformer based framework designed to perform three tasks jointly: ventricular segmentation, physiological metric estimation, and automated disease classification. Unlike prior pipelines that treat these tasks independently, PULSE learns a shared anatomical representation, strengthened through Lov\'asz regularized boundary supervision and volume–wise normalization. Our ablation studies demonstrate that every component contributes measurably to stability: augmentation improves generalization, normalization reduces contour drift, and hybrid loss provides the final gain in myocardial reconstruction fidelity. The resulting model achieved a mean Dice of 0.821 and 81.6\% classification accuracy on ACDC, with clinical index error well within accepted inter observer variability. A core contribution of this work is generalizability. 

Without retraining, PULSE transferred effectively to M\&Ms and
Sunnybrook MRI, maintaining anatomical coherence despite scanner differences. Under extreme modality shift, few shot fine–tuning on CAMUS ultrasound rapidly restored performance (mDice: 0.612 $\rightarrow$ 0.815), demonstrating that strong cardiac priors can be reused even when image appearance and noise characteristics change entirely. This outcome is
particularly promising for hospitals with limited labelled ultrasound data, where rapid deployment is a practical necessity. Qualitative visualizations across five disease cohorts showed clinically meaningful segmentation behaviour , enlarged DCM
ventricles, concentric remodeling in HCM, wall thinning in MINF, and high–variance RV geometry were captured with fidelity. Remaining limitations include occasional myocardium ambiguity in apical slices and sensitivity to heavy echocardiographic noise, indicating room for temporal smoothing, uncertainty modelling, and adaptive post–processing. This study demonstrates that cardiac AI systems can be trained once, transferred broadly, and adapted with minimal
data. PULSE provides a foundation toward unified multi modality cardiac interpretation, by enabling reduced annotation cost and moving closer to deployable, real time clinical decision support. Next steps for the continued refinement of PULSE will include full sequence temporal reasoning, 3D volumetric
consistency, and personalized calibration across diverse patient populations

\section{Data Availability}
This study utilizes four publicly accessible cardiac imaging datasets covering MRI and echocardiography modalities. All datasets are released for research use and were obtained under their respective data usage terms. No proprietary or patient-identifiable clinical data were used. 

Automated Cardiac Diagnosis Challenge : Publicly available CMR dataset with manual LV/RV/Myo annotations. Accessible through the MICCAI ACDC challenge portal upon registration for research use. \url{https://www.creatis.insa-lyon.fr/Challenge/acdc/}

Sunnybrook Cardiac Data (SCD): 
Open cardiac MRI dataset containing healthy and pathological subjects with expert ventricular labels. Distributed by the Sunnybrook Health Sciences Centre under academic use license \url{https://www.cardiacatlas.org/sunnybrook-cardiac-data/}

M\&M-2 (Multi-Centre Multi-Vendor CMR Dataset) :
MRI cine dataset spanning multiple scanners, vendors, and pathological states. Data access is available through the M\&Ms 2021/2023 challenge platform with research agreement approval. \url{https://www.ub.edu/mnms-2/}

CAMUS Echocardiography Dataset: 
Public 2-chamber and 4-chamber ultrasound dataset with ground-truth contours for LV segmentation. Freely accessible for non-commercial scientific use \url{https://www.creatis.insa-lyon.fr/Challenge/camus/}. 

\bibliographystyle{elsarticle-num}

\bibliography{refs}

\end{document}